# Replanning in Domains with Partial Information and Sensing Actions

**Ronen I. Brafman**  BRAFMAN@CS.BGU.AC.IL
*Department of Computer Science*
*Ben-Gurion University of the Negev*

**Guy Shani**  SHANIGU@BGU.AC.IL
*Department of Information Systems Engineering*
*Ben-Gurion University of the Negev*

## Abstract

Replanning via determinization is a recent, popular approach for online planning in MDPs. In this paper we adapt this idea to classical, non-stochastic domains with partial information and sensing actions, presenting a new planner: SDR (*Sample, Determinize, Replan*). At each step we generate a solution plan to a classical planning problem induced by the original problem. We execute this plan as long as it is safe to do so. When this is no longer the case, we replan. The classical planning problem we generate is based on the translation-based approach for conformant planning introduced by Palacios and Geffner. The state of the classical planning problem generated in this approach captures the belief state of the agent in the original problem. Unfortunately, when this method is applied to planning problems with sensing, it yields a non-deterministic planning problem that is typically very large. Our main contribution is the introduction of state sampling techniques for overcoming these two problems. In addition, we introduce a novel, lazy, regression-based method for querying the agent's belief state during run-time. We provide a comprehensive experimental evaluation of the planner, showing that it scales better than the state-of-the-art CLG planner on existing benchmark problems, but also highlighting its weaknesses with new domains. We also discuss its theoretical guarantees.

## 1. Introduction

In many real world scenarios an agent must complete a task where some required features are unknown, but can be observed through special sensing actions. Consider for example a Mars rover that must collect rock samples (Smith & Simmons, 2004). The rover does not know which of the rocks surrounding it contains an interesting mineral, but it can move closer to the rocks and then activate a sensor that detects such minerals. Such domains can be modeled as planning under partially observability with sensing actions (PPOS).

Planning under partial observability with sensing actions is one of the hardest problems for automated planning. Its difficulty stems from the large number of contingencies that can occur, and the need to take them into account while planning. To address these contingencies, the planner must generate a conditional plan, or plan tree, rather than a linear plan. This plan tree can grow exponentially in the number of propositions in the problem description, making offline generation of a complete plan tree impossible for even moderately complex problems. This difficulty can be overcome, to some extent, by using an online planner which generates the next action only, given its current state. One technique for online planning is *replanning* (Zelinsky, 1992), made popular by the FF-replan planner (Yoon, Fern, & Givan, 2007). In replanning, at each state, the agent finds a BRAFMAN & SHANIplan based on a partial, possible inaccurate model, or using a simpler planning problem. It executes some prefix of it, replanning when new information arrives.

The key component of any replanning algorithm is a method for generating and solving such simpler problems. Recent replanners focus on generating fully-deterministic classical planning problems and solving them using off-the-shelf (Yoon et al., 2007), or modified (Albore, Palacios, & Geffner, 2009) classical planners. This approach is particularly beneficial given the large array of existing classical planners, and the ability to immediately enjoy any progress made in this intensively studied area. However, this still leaves open the key question of how to generate an appropriate classical planning problem given the agent's current state. In the context of probabilistic planning with full observability, current planners use multiple samples of classical planning problems obtained by transforming stochastic actions into deterministic actions by selecting a single effect for each action instance (Yoon et al., 2007; Yoon, Fern, Givan, & Kambhampati, 2008; Kolobov, Mausam, & Weld, 2010).

In the context of PPOS a more sophisticated translation scheme was introduced in the CLG planner (Albore et al., 2009). This translation is based on techniques introduced by Palacios and Geffner (2009) for solving conformant planning by representing the agent's belief state within the classical planner's state. This is achieved by extending the language with propositions of the form $Kp$, and $K\neg p$, denoting the fact that the agent "knows" that $p$ is *true* and *false*, respectively. These ideas can be extended to PPOS, but result in non-deterministic planning problems because the effect of a sensing action on an agent's belief state cannot be known offline. CLG handles this problem by further relaxing this problem in a number of ways, using a more complex translation. A key aspect of this translation is that, if $a$ is an action that has a precondition $p$, and the value of $p$ can be sensed, CLG will plan as-if $p$ is true (and thus, $a$ can be executed), provided that an action $a'$ that senses $p$ is executed before $a$. Thus, it makes optimistic assumptions regarding future outcomes of sensing actions. This, however, does not imply that CLG will actually execute $a$, because if the actual outcome of $a'$ differs from the expected one, CLG will replan using its new information.

The SDR planning algorithm we propose in this paper follows a similar high-level approach based on replanning. At each state, we generate a classical planning problem that reflects information about the agent's belief state. The specifics of our approach, are, however a bit different, and our main aim is to provider better scalability, which requires generating smaller classical planning problems. We achieve this by using state sampling. That is, instead of planning for all possible initial states, we sample a small subset of states, and plan as if they are the only possible initial states. We also use sampling to remove the optimistic bias of CLG: we sample an arbitrary initial state $s_I$, and assume that observation values are those obtained when $s_I$ is the true initial state. This use of sampling leads to much smaller classical planing problems, and hence, to better scalability.

Because our planner operates under assumptions that are not alway true (i.e., it considers only a subset of initial states and one possible set of observations), it is possible that the preconditions of actions it selected, as well as the goal, do not hold in all possible worlds. Thus, the agent must maintain some representation of its current set of possible states, also known as the *belief state*, in order to verify these conditions. There are many methods for maintaining a belief state (Albore et al., 2009; To, Pontelli, & Son, 2011; To, Son, & Pontelli, 2011a; To, Pontelli, & Son, 2009), all of which work well for problems with certain structure, and not as well on other problem structures. In general, belief state maintenance is difficult, but because our use of the belief state is limited, we suggest a lazy belief state querying mechanism in the spirit of CFF (Hoffmann & Brafman, 2006) that does not require an explicit representation or update of the belief state following each action. We maintain a symbolic representation of the initial belief state, only. To determine if some literal holds in the current belief state, we regress the literal through the history of actions and observations, and check the consistency of the regressed formula with the initial belief state. We augment the regression process with a caching mechanism, which we call *partially-specified states*, that allows us to keep the regressed formulas compact.

The resulting planner – SDR (*Sample, Determinize, Replan*) – compares favorably with CLG (Albore et al., 2009), which is the current state-of-the-art contingent planner. On most existing benchmarks it generates plans faster and can solve problems that CLG cannot solve, and its plans have similar or slightly worse size.

Our paper contains a comprehensive experimental evaluation aimed at identifying the strengths and especially the weaknesses of SDR and the replanning approach. To this end, we formulated a number of new benchmark domains, including domains where sensing requires performing actions that are off the path to the goal, and domains with dead-ends. In addition, we also evaluate the effectiveness of our new regression-based method for maintaining information on the belief state by comparing it to the closest lazy approach – that of CFF (Hoffmann & Brafman, 2006). Finally, we describe some theoretical guarantees associated with SDR. First, we show that the translation scheme we use is sound and complete whenever the sampled initial state is the true initial state. Then, we show that, under certain assumptions on the connectivity of the domain, SDR with a complete description of the initial belief state will reach the goal, if the goal is reachable.

The paper is organized as follows: In the next section we describe the problem of contingent planning with partial observability and sensing. Then, we describe an idealized version of the SDR planner that ignores efficiency problems that arise when the belief state is large – it uses the entire belief state to generate the classical planning problem. We provide a theoretical analysis of the correctness and convergence properties of this idealized algorithm. Next, we describe the full SDR algorithm. This algorithm uses state sampling to manage the size of the belief state as well as a regression mechanism for querying the belief state. This is followed by an overview and comparison to related work, followed by an empirical evaluation analyzing the strengths and weakness of SDR.

## 2. Problem Definition

We focus on planning problems with partial observability and sensing actions (PPOS). We shall assume that actions are deterministic throughout the paper.

Formally, PPOS problems can be described by a quadruple: $\langle P, A, \varphi_I, G \rangle$, where $P$ is a set of propositions, $A$ is a set of actions, $\varphi_I$ is a propositional formula over $P$ that describes the set of possible initial states, and $G \subset P$ is the set of goal propositions. In what follows we will often abuse notation and treat sets of literals as a conjunction of the literals in the set, as well as an assignment of values to propositions appearing in this set. For example, $\{p, \neg q\}$ is also treated as $p \wedge \neg q$ and as an assignment of *true* to $p$ and *false* to $q$.

A state of the world, $s$, assigns a truth value to all elements of $P$, and is usually represented using the closed-world assumption via the set of propositions assigned *true* in $s$. A *belief-state* is a set of possible states, and the initial belief state, $b_I = \{s : s \models \varphi_I\}$ defines the set of states that are possible initially.

An action, $a \in A$, is a three-tuple, $\{pre(a), effects(a), obs(a)\}$. The action preconditions, $pre(a)$, is a set of literals that must be valid before the action can be executed. The action effects, *effects*(a), is a set of pairs, $(c, e)$, denoting conditional effects, where $c$ is a set (conjunction) of literals and $e$ is a single literal.

Finally, *obs*$(a)$ is a set of propositions, denoting those propositions whose value is observed following the execution of $a$. We assume that $a$ is consistent, that is, if $(c, e) \in$ *effects*$(a)$ then $c \wedge pre(a)$ is consistent, and that if both $(c, e), (c', e') \in$ *effects*$(a)$ and $s \models c \wedge c'$ for some state $s$, then $e \wedge e'$ is consistent.

In current benchmark problems, either the set *effects* or the set *obs* are empty. That is, actions either alter the state of the world but provide no information, or they are pure sensing actions that do not alter the state of the world, but there is no reason for this to be the case in general.

We use $a(s)$ to denote the state that is obtained when $a$ is executed in state $s$. If $s$ does not satisfy all literals in *pre*$(a)$ then $a(s)$ is undefined. Otherwise, $a(s)$ assigns to each proposition $p$ the same value as $s$, unless there exists a pair $(c, e) \in$ *effects*$(a)$ such that $s \models c$ and $e$ assigns $p$ a different value than $s$. If $\bar{a}$ is a sequence of actions, we use $\bar{a}(s)$ to denote the resulting state, defined analogously.

Observations affect the agent's belief state. We assume that all observations are deterministic and accurate, and reflect the state of the world *prior* to the execution of the action.[1] Thus, if $p \in$ *obs*$(a)$ then the agent will observe $p$ if $p$ holds now (i.e., prior to $a$'s effect), and otherwise it will observe $\neg p$. Thus, if $s$ is the true state of the world, and $b$ is the current belief state of the agent, then $b_{a,s}$, the belief state following the execution of $a$ in state $s$ is defined as:

$$b_{a,s} = \{a(s') | s' \in b, s' \text{ and } s \text{ agree on } obs(a)\}$$

That is, the progression through $a$ of all states where the agent would receive, following the execution of $a$, the same observation as if it was at state $s$. Extending the definition to sequences of actions, if $\bar{a}$ is a sequence of actions, then $b_{\bar{a},s}$ denotes the belief state reached from $b$ when executing $\bar{a}$ starting at state $s \in b$.

Alternatively, we can define the belief progression without an explicit world state $s$ using:

$$b_{a,o} = \{a(s') | s' \in b, \text{ and the observation of } a \text{ when } s' \text{ is the world state is } o\}.$$

A contingent plan for a PPOS problem is an annotated tree $\tau = (N, E)$. The nodes, $N$, are labeled with actions, and the edges, $E$, are labeled with observations. A node labeled by an action with no observations has a single child, and the edge leading to it is labeled by the null observation *true*. Otherwise, each node has one child for each possible observation value, i.e., one child for each possible assignment to the observed propositions. The edge leading to this child is labeled by the corresponding observation. The plan is executed as follows: the action at the root is executed, an observation (possibly null) is made, and the execution continues recursively from the child that corresponds to the edge labeled with this observation. As we assume that actions and observations are deterministic, there is a single possible execution path along this tree for each initial state. We use $\tau(s)$ to denote the state obtained when $\tau$ is executed starting in state $s$. $\tau$ is a *solution plan* (*plan* for short) for PPOS $\mathcal{P} = \langle P, A, \varphi_I, G \rangle$ if $\tau(s) \models G$ for every $s \models \varphi_I$.

Complete contingent plans that consider all possible future observations, can be prohibitively large for problems of practical interest. In this paper, thus, we are concerned with online planning in PPOS. That is, at each stage, the planner selects only the next action to execute, executes it, and reconsiders.

---

1. One can choose to have the observations reflect the state of the world following the execution of the action at the price of a slightly more complicated notation below.

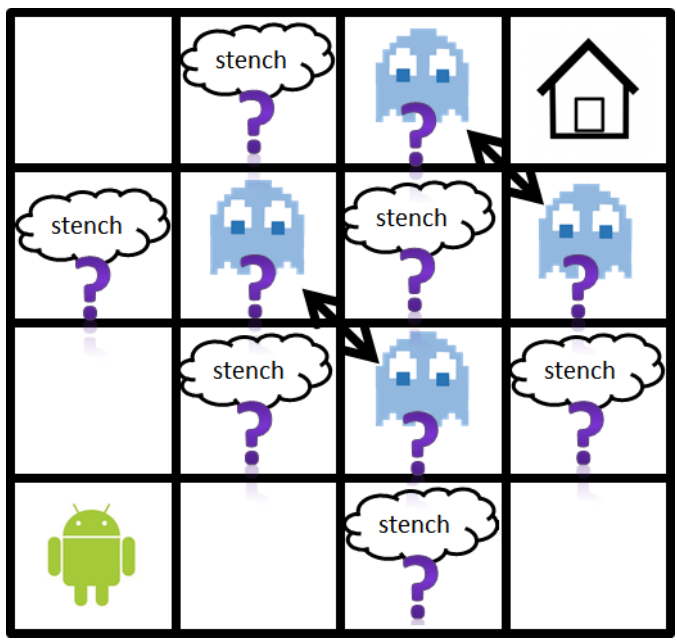

**Figure 1:** The $4 \times 4$ Wumpus Domain

*Example* 1. We illustrate these definitions using a $4 \times 4$ simplified Wumpus domain (Albore et al., 2009), which will serve as our running example. In this domain, illustrated in Figure 1, an agent navigates on a $4 \times 4$ grid from the bottom-left corner to the top-right corner (the goal) by moving in any of the four directions. The squares around the top two squares of the diagonal of this grid contain monsters called Wumpus – one for every pair of squares adjacent to the diagonal (e.g., there's either a Wumpus in square 3,4 or in 4,3). An agent can only move into a square if it is *safe*. That is, if it contains no Wumpus, specified as a precondition to the move action. Thus, the agent can never enter a square with a Wumpus and die, and there are no dead-ends in this domain.[2] At the initial state the agent does not know the location of the Wumpuses, nor can it directly observe their location. However, a Wumpus emits a stench that drifts to all the adjacent squares. Hence, when the agent is at an adjacent square to a Wumpus it can smell the stench, although it cannot determine on which adjacent square the Wumpus is hiding. Thus, measurements in a few different locations may be required to determine the precise position of a Wumpus. This domain demonstrates a complex hidden state with multiple sensing actions, but with no conditional effects.

We formalize this problem as follows:

- The set of propositions is *at-x-y* for $1 \leq x, y \leq 4$, *wumpus-at-2-3*, *wumpus-at-3-2*, *wumpus-at-3-4*, *wumpus-at-4-3* and *stench-at-x-y* for $1 \leq x, y \leq 4$.

- The actions are *move-from-$x_1$-$y_1$-to-$x_2$-$y_2$* for all adjacent $\langle x_1, y_1 \rangle, \langle x_2, y_2 \rangle$ pairs, and *smell*.

- The initial state is: *at-1-1* $\wedge \neg$ *at-1-2* $\wedge \cdots \wedge \neg$ *at-4-4* $\wedge$ (*oneof wumpus-at-2-3 wumpus-at-3-2*) $\wedge$ (*oneof wumpus-at-4-3 wumpus-at-3-4*).

---

2. Later on, we introduce a more natural formalization of this domain that does not require a square to be safe to move into it, and thus, contains dead-ends.

- The goal is *at-4-4*.

- The initial belief state consists of four possible world-states, corresponding to the four possible truth assignments to (*wumpus-at-2-3* ∨ *wumpus-at-3-2*) ∧ (*wumpus-at-4-3* ∨ *wumpus-at-3-4*) in addition to the known literals, such as *at-1-1* and the adjacency propositions.

## 3. The Idealized SDR Planner

We now describe an idealized version of the Sample Determinize Replan (SDR) planner. It samples a single distinguished state $s'$ from the current belief, and creates a deterministic classical problem, where observations correspond to those derived from initial state $s'$. After solving the classical problem using any classical planner, it applies the resulting plan until some sensing action is performed. Then, the belief state is updated, removing all states that do not agree on the observed value, and the process is repeated. In this idealized version of SDR a complete and explicit description of the agent's belief state is maintained and used to generate the classical planning problem. The actual algorithm modifies this version by using a sampled belief state and lazy belief-state maintenance; it is described in Section 5.

### 3.1 Replanning using Complete Translations

The central component of SDR is a translation from PPOS to a classical planning problem. It is well known that planning under uncertainty can be reduced to planning in belief space, i.e., where we move between sets of possible states of the world, modeling our current knowledge. When there are no sensing actions, as in conformant planning, this results in a classical, deterministic planning problem in belief space. When sensing actions exist, however, the resulting problem is non-deterministic, because the effect of sensing actions on the agent's state of knowledge depends on the values observed online, which depend on the true state of the world, hidden from the agent.

Since we want to generate a deterministic classical planning problem that can be solved by an off-the-shelf classical planner, we need to determinize the non-deterministic problem described above, simplifying it in the process. We do this by selecting one possible initial state $s'$ and assuming that observation values correspond to those obtained when $s'$ is the true initial state. Note that this does not imply that the planner plans as if $s'$ was the actual initial state of the world (which would yield a very simple classical planning problem over a standard state space) because the planner actually reasons about its belief state online when $s'$ is the true initial state, and attempts to reach a belief state in which the goal is known, i.e., where all possible states are goal states.

In most cases, the assumption that $s'$ is the true initial state turns out to be incorrect. The planner learns this online when it executes a sensing action and discovers that its outcome was different from what is expected if $s'$ was the true initial world state. At this point, we have learned that $s'$ (and possibly some other initial states) cannot possibly be the true initial world state. Thus, our uncertainty is reduced, and we replan using the new belief state and a new initial state sampled from it.

The algorithm terminates when we cannot find a solution to the classical problem generated, which implies that the goal cannot be achieved from the set of states indistinguishable from the selected state $s'$, or, when our belief state indicates that the goal, $G$ is known, i.e., for every state $s \in b$, $s \models G$.

The high-level SDR algorithm with a complete initial belief state is described in Algorithm 1.

**Algorithm 1** SDR (Complete Translation)

**Input:** PPOS Problem: $\mathcal{P} = \langle P, A, \varphi_I, G \rangle$, Integer: *size*
1: $b^0 :=$ the initial belief state $b_I = \{s : s \models \varphi_I\}$
2: $i := 0$
3: **while** $G$ does not hold in all states in $b^i$ **do**
4:     Select a state $s' \in b^i$
5:     Generate a deterministic planning problem $C$ given $\mathcal{P}, b^i, s'$
6:     Find a solution plan $\pi$ for $C$
7:     **if** no solution exists **then**
8:         **return** failure
9:     **end if**
10:     **while** $\pi \neq \emptyset$ **do**
11:         $a :=$first$(\pi)$
12:         Execute $a$, observe $o$
13:         $b^{i+1} \leftarrow b^i_{a,o}$ – update the belief given $a, o$
14:         $i \leftarrow i + 1$
15:         Remove $a$ from $\pi$
16:         **if** $o$ is inconsistent with $s'$ **then**
17:             break
18:         **end if**
19:     **end while**
20: **end while**

### 3.2 Generating a Classical Planning Problem

Given the input PPOS $\mathcal{P} = \langle P, A, \varphi_I, G \rangle$, the current belief state $b$, and the selected state $s' \in b$ (hypothesized to be the current true system state), we generate a classical planning problem $\mathcal{P}_c(b, s') = \langle P_c(b), A_c(b), I_c(b, s'), G_c \rangle$. Notice that $s'$ influences the definition of the classical initial state only, while $b$ influences all elements except for the goal. $\mathcal{P}_c(b, s')$ is defined as follows:

**Propositions** $P_c(b) = P \cup \{Kp, K\neg p | p \in P\} \cup \{p/s | p \in P, s \in b\} \cup \{K\neg s | s \in b\}$ :

1. $P$ – The set of propositions that appear in the original problem. Their value is initialized according to the distinguished state $s'$ and are updated to reflect the current state of the world given that $s'$ is the true initial state.

2. $\{Kp, K\neg p | p \in P\}$ – Propositions encoding knowledge obtained by the agent. $Kp$ holds if the agent knows $p$ is true, i.e., if $p$ holds in all possible states. This knowledge can be obtained through a sensing action in which $p$ was observed to be true or as a necessary consequence of an action. The agent can know that $p$ is true (denoted $Kp$), know that $p$ is false ($K\neg p$), or not know the value of $p$ (denoted by both $\neg Kp$ and $\neg K\neg p$).

3. $\{p/s | p \in P, s \in b\}$ – Propositions that capture the value of $p$ given that $s$ is the true initial state. We can use them to rule out certain states. For example, if we observed $p$ to be true, we can rule out any state $s$ as the true initial state if $\neg p/s$ holds.

4. $\{K\neg s | s \in b\}$ – Propositions that capture which states have been ruled out. When concluding that a certain state $s$ was not the initial state of the system, we acquire $K\neg s$.

Below we shall use $Kc$ as a shorthand notation for $Kl_1 \wedge \cdots \wedge Kl_m$, where $c = l_1 \wedge \cdots \wedge l_m$, and $\neg K\neg c$ as a shorthand notation for $\neg K\neg l_1 \wedge \cdots \wedge \neg K\neg l_m$. We also note that the number of propositions generated in the idealized translation can be exponentially large, and this is why the actual SDR planner, described in Section 5, uses various approximations.

**Actions** For every action $a \in A$, $A_c(b)$ contains an action $a_c$ defined as follows:

*pre*($a_c$) = *pre*($a$) $\cup \{Kp | p \in pre(a)\}$. That is, the precondition of the action must hold *and* the agent must *know* this to be true prior to applying the action.

For every $(c, e) \in$ *effects*($a$), *effects*($a_c$) contains the following conditional effects:

1. $(c, e)$ – the original effect. These conditions update the state that is assumed to be the true state of the world.

2. $\{(c/s, e/s) | s \in b\}$ – the above, conditioned on the states consistent with $b$. These conditions update the values of propositions given the possible states in $b$.

3. $(Kc, Ke)$ – if we know that the condition $c$ holds prior to executing $a$, we know that its effect holds following $a$. This condition allows us to gain knowledge.

4. $(\neg K\neg c, \neg K\neg e)$ – if $c$ is not known to be false prior to executing $a$, $e$ will not be known to be false afterwards.

5. $\{(p, Kp), (\neg p, K\neg p) | p \in obs(a)\}$ – when observing the value of $p$, we gain knowledge in the form of either $Kp$ or $K\neg p$, depending on the value of $p$ in the state that is assumed to be the true world state.

6. $\{(p \wedge \neg p/s, K\neg s), (\neg p \wedge p/s, K\neg s) | p \in obs(a), s \in b\}$ – rule out possible states in $b$ if they are inconsistent with the observation. This is sometimes known as the *refutation* of states.

In addition, for each literal $l$ (w.r.t. $P$) we have a merge action that allows us to conclude absolute knowledge from knowledge conditional on the states in $b$. That is, if all states agree on the value of some proposition, then we know the value of that proposition. We exclude states that were already refuted, i.e., found to be inconsistent with some observation.

- *pre*(merge($l$)) = $\{l/s \vee K\neg s | s \in b\}$ – for each state $s$, either it agrees on $l$, or it was previously refuted.

- *effects*(merge($l$)) = $\{(\text{true}, Kl)\}$.

**Initial State** $I_c(b, s') = \bigwedge_{l:s' \models l} l \ \bigwedge_{p \in P} \neg Kp \wedge \neg K\neg p \ \bigwedge_{s \in b, s \models l} l/s \ \bigwedge_{s \in b} \neg K\neg s$:

1. $\{l : s' \models l\}$ – The set of propositions $P$ that appear in the original problem, initialized to their value in the distinguished state $s'$. Notice that this is the only element of the translation affected by the choice of the hypothesized initial state.

2. $\{\neg Kp \wedge \neg K\neg p : p \in P\}$ – Knowledge propositions $Kp$ and $K\neg p$ are initialized to false, denoting no initial knowledge over the value of propositions.

3. $\{l/s : s \in b, s \models l\}$ – Propositions of the form $p/s$ are initialized to the value of $p$ in the corresponding state $s$.

4. $\{\neg K \neg s : s \in b\}$ – Propositions $K \neg s$ are initialized to false, as we do not know any initial state in $b$ to be impossible before observing the value of some proposition.

**Goal** $G_c = KG$, that is, all goal literals are known to be true.

The above translation is similar to, and inspired by the $K_{S_0}$ translation introduced by Palacios and Geffner (2009) for conformant planning. To adapt it to PPOS, we chose to determinize observation actions by sampling a distinguished initial state, $s'$, whose value we track throughout the plan, and use to select the outcome of observation actions. Albore et al. (2009) provide a different translation that contains propositions that encode the fact that the value of some proposition was sensed (denoted $Ap$). This does not imply that $p$ is true, but that the agent knows the value of $p$. Instead of requiring $Kp$ to hold prior to executing an action with precondition $p$, they require $Ap$ to hold. This results in an optimistic choice of values for sensed propositions.

A natural extension of the idea of conditioning the value of propositions on the initial state, as capture by the $p/s$ propositions we use, it to condition their value on multiple initial states from which $p$ progresses identically. That is, suppose that $s$ and $s'$ differ only on the value of some proposition $r$, and that $r$ does not appear in actions that affect the value of $p$. In that case, $p/s$ and $p/s'$ will always have the same value. Consequently, we can simply maintain a single proposition, $p/\{s, s'\}$. This, indeed, is the approach taken by Palacios and Geffner (2009) and Albore et al. (2009), and such sets of states are called *tags*. The larger the set of states denoted by a tag, the fewer tags needed, and the smaller the representation. In fact, Palacios and Geffner show that most conformant planning problems require a very small set of tags, linear in the number of proposition. The use of tags is an important optimization, that can lead to an exponential reduction in the size of the generated planning problem. We decided not to introduce tags here as this would further complicate the description of translation, and because our primary technique for reducing problem size is state sampling – discussed in Section 5. Nevertheless, SDR could be further optimized by using tags instead of states.

*Example* 2. We now demonstrate the above translation using a small toy example of identifying and treating a disease. There are $n$ possible diseases, each of which can be uniquely identified using a single test, and cured using a unique treatment. Before applying the treatment, we must identify the disease, so as to avoid applying the wrong treatment, causing further damage. The PPOS is hence defined as follows:

- We have one proposition per disease, $disease_i$ for $i \in \{1..n\}$, and a proposition for the result of the test *test-passed*.

- We need $n$ test actions $test_i$, with no preconditions and conditional effect ($disease_i$,*test-passed*), and $n$ treatment actions $treat_i$ with precondition $disease_i$ and effect $\neg disease_i$. We also have one sensing action *observe-test-result* allowing us to sense the value of *test-passed*.

- The initial state is: (*oneof* $disease_1, ..., disease_n$) $\wedge \neg$*test-passed*. The initial belief state consists of $n$ possible world-states, corresponding to the $n$ possible diseases.

- The goal is $\bigwedge_{i \in [1,n]} \neg disease_i$.

We will denote the possible states using $s_i$ where $i \in \{0, 1..n\}$, such that in $s_i$ the patient has disease $i$ and in state $s_0$ the patient has no disease. Let us choose $s_k$ as our $s'$, that is, we choose to assume that the patient has the $k^{th}$ disease.

The set of propositions in the translation is:

- The original propositions $disease_i$ and *test-passed*.

- Propositions representing unconditional knowledge:
    - *Kdisease$_i$*, *K¬disease$_i$* for $i \in \{1..n\}$.
    - *Ktest-passed*, *K¬test-passed*.

- Propositions representing knowledge conditional upon the initial states:
    - $disease_i/s_j$, for $i \in \{1..n\}, j \in \{0..n\}$.
    - *test-passed*$/s_j$, for $j \in \{0..n\}$.

- $K\neg s_j$ for $j \in \{0..n\}$.

The set of actions is:

- Test: for each $test_i$ action we have:
    - No preconditions.
    - effects:
        * ($disease_i$, *test-passed*).
        * ($\neg disease_i$, $\neg$*test-passed*).
        * ($disease_i/s_j$, *test-passed*$/s_j$), for $j \in \{0..n\}$.
        * ($\neg disease_i/s_j$, $\neg$*test-passed*$/s_j$), for $j \in \{0..n\}$.

- Treat: for each $treat_i$ action we have:
    - precondition: $disease_i$
    - effects: $\neg disease_i \wedge K\neg disease_i$.

- Observing the test result:
    - No preconditions:
    - Effects:
        * (*test-passed*, *Ktest-passed*) – positive observations.
        * (*test-passed* $\wedge \neg$*test-passed*$/s_j$), $K\neg s_j$, for $j \in \{0..n\}$ – refuting states that do not agree with the positive observation.
        * ($\neg$*test-passed*, *K¬test-passed*) – negative observations.
        * ($\neg$*test-passed* $\wedge$ *test-passed*$/s_j$, $K\neg s_j$), for $j \in \{0..n\}$ – refuting states that do not agree with the negative observation.

- Merge: we need merges only for the $disease_i$ proposition (see the implementation note below):

- preconditions: $\bigwedge_{j\in[0..n]} disease_i/s_j \vee K\neg s_j$
- effects: $K disease_i$

- and actions for merging $\neg disease_i$:

  - preconditions: $\bigwedge_{j\in[0..n]} \neg disease_i/s_j \vee K\neg s_j$
  - effects: $K\neg disease_i$

The initial state is a conjunction of the following conjuncts (we omit negations for simplicity of presentation):

- The distinguished initial state: $disease_k$

- Uncertainty about the current state: $\neg K\neg s_j$, for $j \in \{0..n\}$.

- Conditional knowledge: $disease_i/s_i$, $\neg disease_i/s_j$, for $i \in \{1..n\}, j \in [0..n], i \neq j$.

Finally, the goal is $\bigwedge_{i\in[1..n]} K\neg disease_i$.

### 3.3 Notes on Efficient Implementation

While the above translation is correct, some of the propositions can be removed. In many problems, some of the original propositions are always known. For example, in the Wumpus example, the location of the agent is always known, and it is identical in all possible current states. If the value of $p$ is always known, we can remove the propositions $Kp$, $K\neg p$, $p/s$, and use proposition $p$ only. These propositions do not require merge actions, either.

The refutation effect of actions (item 6 in the list of action effects above) can be moved out to an independent action, similar to the merge actions. We create for each proposition $p$ that is observable by some action $a$ (i.e., $p \in obs(a)$ for some $a \in A$) and each state $s$ two refutation actions, with preconditions $p \wedge \neg p/s$ or $\neg p \wedge p/s$, and an identical effect $K\neg s$. This reduces the number of conditions in each action which poses a difficulty for current classical planners.

## 4. Theoretical Guarantees

We now prove two important properties of our algorithm when applied to a deterministic PPOS $\mathcal{P}$. Our first result is a proof of the correctness of the translation. We show that a plan exists for the original problem iff a plan exists for the classical problems we generate, assuming we guessed the correct initial state. We then show that, under standard assumptions, our algorithm will eventually reach the goal. We note that an understanding of these results is not required for the following sections.

We begin this section with some definitions of notations that will be used in the theorems below:

- **Classical planning notations**: if $b$ is a belief state and $s \in b$ is a (world) state then recall that $I_c(b, s)$ denotes the initial state of the classical planning problem SDR's translation would generate when $s$ is selected as distinguished initial state. If $\pi$ is a plan for $\mathcal{P}$ and $b$ is the initial belief state then $\pi_c$ is the corresponding plan for the classical planning problem $\mathcal{P}_c(b, s)$, where each action $a$ in $\pi$ is replaced by the corresponding action in the generated classical problem. In addition, prior to the first action of $\pi_c$, and following all the translated actions, all *merge* actions are inserted. We will assume a modified version of the *merge* actions that has no preconditions and instead, its current precondition replaces the condition part of the conditional effect which was previously empty (i.e., = *true*). This allows us to insert arbitrary *merge* actions without risking making the plan undefined. Adding all merges, the plan $\pi_c$ is forced to make all possible inferences following every action.

- **Sensing and non-sensing actions**: without loss of generality, we will assume that each action either makes an observation or changes the state of the world, but not both. Actions that do both can be modeled as a consecutive pair of actions; first, one that changes the world, and then one that makes observations.

- **Indistinguishable states**: we say that $s, s'$ are *indistinguishable* by $\pi$ if $\pi$ is applicable to $s$ iff it is applicable to $s'$, and the observations generated when executing $\pi$ from $s$ and $s'$ are identical. Note that $s, s'$ may be indistinguishable by $\pi$ but distinguishable by some other plan $\pi'$.

- **Applicability of actions and plans**: we say that an action $a$ is *applicable* in state $s$ if $s$ satisfies all the preconditions of $a$. We say that an action $a$ is applicable in a belief state $b$ if each $s \in b$ satisfies all the preconditions of $a$. Applicability is generalized to plans in the natural manner. That is, given a plan $\pi = \langle a_1, ..., a_n \rangle$ and a state $s$, the plan is applicable for $s$ if $a_1$ is applicable to $s$ and $\pi' = \langle a_2, ..., a_n \rangle$ is applicable to $a_1(s)$.

We begin by showing that if $\pi_c$ achieves $Kl$ for some literal $l$, then the belief resulting from executing the plan $\pi$ in belief space will satisfy $l$.

**Theorem 1.** *Let $b$ be a belief state and $s \in b$ the true initial state. Let $\pi$ be a sequence of actions in the original problem. Then for every literal $l$, $b_{\pi,s} \models l$ iff $\pi_c(I_c(b,s)) \models Kl$.*

*Proof.* We prove by induction on $|\pi|$ that the following conditions hold:

1. $\pi$ is applicable in $b$ and $s$ iff $\pi_c$ is applicable on $I_c(b,s)$

2. For every $s' \in b$ that is indistinguishable from $s$ by $\pi$ and every literal $l$: $\pi(s') \models l$ iff $\pi_c(I_c(b,s)) \models l/s'$

3. For every $s' \in b$, $s'$ is distinguishable from $s$ by $\pi$ iff $\pi_c(I_c(b,s)) \models K\neg s'$

4. For every literal $l$: $b_{\pi,s} \models l$ iff $\pi_c(I_c(b,s)) \models Kl$

**Base case.** In the base case $\pi = \emptyset$ and $\pi_c$ includes all merge actions. Conditions 1, 2, and 3 are immediate by construction. That is, the empty plan is always applicable (because it contains no actions), and there are no distinguishable states. Condition 4 is a consequence of the definition of the *merge* action.

**Inductive step.** For the inductive step, we assume that conditions 1-4 hold for $\pi'$ and consider a sequence $\pi = \pi' \cdot a$. We consider two cases:

- $a$ **is not a sensing action**: For condition 1, observe that given the induction hypothesis, we need only show that $a$ is applicable following $\pi'$ iff $a_c$ is applicable following $\pi'_c$. (If some prefix is inapplicable in one case, we know by the induction hypothesis that it is also inapplicable in the other). Suppose that $a$ is applicable, i.e., $b_{\pi',s} \models p$ for every precondition $p$ of $a$. From condition 4 we conclude that $\pi'_c(I_c(b,s)) \models Kp$ which is the corresponding precondition of $a_c$. The additional merge actions have no preconditions. The other direction is similar.

For conditions 2 and 3 notice that indistinguishable states become distinguishable only by observing some literal that holds in one but not in the other. Thus, a non-sensing action $a$ cannot cause two states that were indistinguishable given $\pi'$ to become distinguishable following $\pi$. Hence, Condition 3 follows immediately from the induction hypothesis and the fact that a non-sensing action does not have any effect of the form $K \neg s$. For Condition 2: If $l$ was not affected by the last action in $\pi$, then this follows from the induction hypothesis. Otherwise, $l$ is added by $a$ under some condition $c$ that holds in $s'$. By construction, $(c/s', l/s')$ is a conditional effect of $a_c$. Given the induction hypothesis, $c$ held prior to the execution of $a$ iff $c/s'$ held prior to the execution of $a_c$ and all the merge actions that follow $\pi'_c(I_c(b,s))$ in $\pi_c(I_c(b,s))$. Consequently, $l/s'$ holds if $l$ holds.

Finally, for Condition 4, $b_{\pi,s} \models l$ iff for every $s'$ that is indistinguishable from $s$ given $\pi$, $\pi(s') \models l$. From Condition 2 above, this happens iff $\pi_c(I_c(b,s)) \models l/s'$ for every such $s'$. Condition 3 guarantees that $\pi_c(I_c(b,s)) \models K \neg s''$ for every $s''$ that is distinguishable from $s$. Thus, a suitable merge action, included by definition in $\pi_c$ will conclude $Kl$. For the other direction, if $Kl$ holds, we know that this was either true before and not affected by any action, or a consequence of a merge action. In the former case, we know using the induction hypothesis that $b_{\pi',s} \models l$. Since $l$ was not influenced by $a$, we conclude that $b_{\pi,s} \models l$. In the latter case of a merge action, we use Condition 2 and 3 to conclude that $b_{\pi,s} \models l$.

- **$a$ is a sensing action**: For states $s'$ that remain indistinguishable from $s$, conditions 1,2 are immediate: the only effect of $a$ is to deduce $Kl$ for some literal $l$. For Condition 1 note that sensing actions are always applicable. For Condition 2 note that sensing actions do not affect the state.

For Condition 3; if $s'$ is distinguishable from $s$ given $\pi$ then either it was distinguishable before, in which case it is distinguishable following $a$, because $K \neg s'$ is never removed once deduced ($\neg K \neg s'$ is not an effect of any action in the translation). If $s'$ just became distinguishable after $a$ is executed, then by construction $a_c$ has $K \neg s'$ as an effect. For the other direction, if $s'$ is indistinguishable from $s$ given $\pi$, it must have been indistinguishable given $\pi'$. Hence, by the induction hypothesis, $\pi'_c(I_c(b,s)) \not\models K \neg s'$. Since $s$ and $s'$ are indistinguishable now, the sensing action has the same effect for $s$ and $s'$, and so $a_c$ does not add $K \neg s'$, by construction.

For Condition 4, first suppose that $b_{\pi',s} \models l$. From the induction hypothesis this happens iff $\pi'_c(I_c(b,s)) \models Kl$. However, $b_{\pi',s} \models l$ and $\pi'_c(I_c(b,s)) \models Kl$ are not affected by $a$, and thus remain true under $\pi$ as well. Therefore, we need consider only the case that $b_{\pi',s} \not\models l$ and $\pi'_c(I_c(b,s)) \models \neg Kl$.

Suppose that $b_{\pi',s} \not\models l$ and $b_{\pi,s} \models l$. This implies that all state $s' \in b$ such that $\pi(s') \models \neg l$ are distinguishable from $s$ by $\pi$. Thus, we have that $\pi_c(I_c(b,s)) \models K \neg s'$ for all such $s'$ and using the merge action, we conclude $\pi_c(I_c(b,s)) \models Kl$.

Next, suppose $b_{\pi',s} \not\models l$ and $b_{\pi,s} \not\models l$. Thus, there exists some $s' \in b$ that is indistinguishable from $s$ given $\pi$ such that $\pi(s') \models \neg l$. Since $b_{\pi',s} \not\models l$ and no merge action can be applied, we conclude that $b_{\pi,s} \not\models l$.

$\square$

This result provides a local soundness and completeness proof (i.e., per replanning phase). Applying the theorem to the goal $G$ we get the following corollary:

**Corollary 1.** *Let $b$ be a belief state, $s$ a state in it, and $G$ the goal. SDR (with a sound and complete underlying classical planner) will find a (standard, sequential) plan that when executed from $s$ with initial belief state $b$ will reach a belief state satisfying $G$, iff one exists.*

*Proof.* From Theorem 1 we have that $b_{\pi,s} \models l$ iff $\pi_c(I_c(b,s)) \models Kl$ for any literal $l$. Thus, $b_{\pi,s} \models G$ iff $\pi_c(I_c(b,s)) \models KG$. Thus, a plan exists for the original problem iff a plan exists for the translated problem. (Recall that $KG$ is a shorthand for $Kg_1 \wedge \cdots \wedge Kg_m$, where $g_i$ are literals and $G = g_1 \wedge \cdots \wedge g_m$.) $\square$

We can also use the theorem to deduce that the belief state is reduced in every replanning episode:

**Corollary 2.** *Let $\pi$ be a plan generated by SDR (with a sound and complete underlying classical planner) for belief state $b$ and initial state $s \in b$. Let $s'$ be the real initial state. If we execute $\pi$ from $b$ and we do not reach a belief state satisfying $G$, then $s' \neq s$.*

*Proof.* Given Theorem 1, we know that SDR will generate a correct plan for $b$ and $s$. Thus, if $\pi$ does not reach the goal then $s$ and $s'$ are not indistinguishable, and consequently, they must be different. $\square$

What we show next is a more global version of correctness. We will show that under standard, but admittedly strong assumptions, our algorithm will reach the goal within a finite number of steps. As above, we assume that a complete representation of the belief state is maintained (as opposed to using a sample, as we do later).

Dead-ends are a well known pitfall for replanning algorithms, but also for most online algorithms that do not generate a complete solution to the problem, which is unlikely to be feasible in practice, and may require exponential space. Thus, to provide reasonable guarantees, our analysis focuses on domains with no dead-ends. More formally, we say that a PPOS problem defines a *connected* state-space, or is *connected*[3], if a state satisfying $G$ is reachable from every state $s$ reachable from $I$.

Another problem in the case of partial observability is the inability to recognize that the goal has been reached. Consider the following simple example; there is a single proposition $p$, which is unobservable and whose value is unknown in the initial state. We have a single action, *flip*, which flips its value. The goal is $p$, and evidently it can be reached from any state. Yet, we can never know that $p$ holds. That is, we can never reach a belief state in which all states satisfy $p$.

Thus, having no deadends is not enough, and we must also require no "belief dead-ends", i.e., belief states from which a belief state satisfying the goal is not reachable. We say that a PPOS is *belief-connected* if given any belief state $b$ reachable from the initial belief state, $b_I$, and any world-state $s$ consistent with $b$ (i.e., $s \in b$), there exists a sequence of actions $\bar{a}$ that, when $s$ is the true starting state, leads from $b$ to $b'$, such that $b' \models G$. In notation: $b_{\bar{a},s} \models G$. Clearly, belief-connectedness implies connectedness.

---

3. The analogous term in the context of RL algorithms and MDPs is *ergodic*.

**Theorem 2.** *Given a belief-connected PPOS problem, SDR with a sound and complete underlying classical planner and a set of tags that corresponds to all possible initial states, will reach the goal after a finite number of steps.*

*Proof.* The proof is based on ideas from the PAC-RL algorithms $E^3$ (Kearns & Singh, 2002) and $R_{max}$ (Brafman & Tennenholtz, 2003) and follows immediately from Corrolary 2. Consider a plan $\pi$ generated under the assumption that $s$ is the true initial state. This plan will be successful from every initial state $s'$ that is indistinguishable from $s$ given $\pi$. If the plan fails, we will recognize this fact because we maintain a sound and complete description of the current belief state. We will also conclude that neither $s$ nor any other initial state indistinguishable from $s$ given $\pi$ is possible, and hence, our belief state will be reduced by at least one state. Because of belief-connectedness, we can still reach the goal. We can continue this process at most a number of times that is equal to the size of the initial belief state, at which point our belief state is a singleton, and we are left with a classical planning problem, that our underlying classical planner will solve. □

The above proof makes it apparent that in many cases, only a polynomial number of replanning phases is required, because the number of initial states ruled out could be very large, e.g., if in each iteration we learn the initial value of a single proposition. Bonet and Geffner (2011) identify one such case, which, consequently, requires only a linear number of replanning phases to succeed.

## 5. SDR with Belief State Sampling and Belief Tracking

The size of the classical problem generated by the translation method suggested in Section 3.2 depends on the cardinality of the belief state, i.e., the number of possible initial states: We generate one proposition for each original proposition and possible initial state, and for each conditional effect of every action we generate a conditional effect for each possible initial state. This can lead to an exponential (in the size of $\mathcal{P}$) larger classical planning problem. Thus, the generated problem's size may become too large for current classical solvers to handle. In addition, the belief state itself can be exponentially large, and an explicit representation of it is impractical. We address these issues by using only a sampled subset of the current belief state to generate the classical planning problem, and by using an implicit description of the belief state.

### 5.1 Sampling the Belief State

To address the problem of a large belief state we suggest using a sampled subset of the belief state to generate the classical planning problem. Conceptually and technically, the change required to the method described in Section 3.2 is minor: select a subset $S'$ of $b$ and generate the classical planning problem as if $S'$ is the true belief state. This also implies that the distinguished initial state $s'$ is chosen from $S'$. Thus given a PPOS $\mathcal{P} = \langle P, A, \varphi_I, G \rangle$, the sampled set of state $S'$ such that $S' \models \varphi_I$, and the distinguished initial state $s' \in S$, we simply generate the classical planning problem $\mathcal{P}_c(S', s')$.

While the sampling translation method is similar to the complete translation, there is an important semantic difference. While in the complete translation $Kp$ denoted *knowing* that $p$ is true in *all possible states*, in the sampling translation $Kp$ denotes knowing that $p$ is true only in *all sampled states*. Thus, upon execution, when the agent intends to execute an action, there might be some preconditions whose value is not known to be true in all possible states. We call such actions *unsafe*. We must ensure that unsafe actions are never executed.

The new translation that uses $S'$ instead of $b$ cannot guarantee that no action in the resulting plan is unsafe. For this reason, we must maintain a representation of the true belief state throughout execution. We use this information to check whether some possible state exists for which some precondition of the next action does not hold. We call such state a *witness state*. If a witness state is found, we must sample again and replan. To ensure that we do not generate another plan that is not executable from the witness state, we add it to the set of sampled states, $S'$. The new plan will either learn to distinguish between the witness state and the rest of the states in $S'$, or choose a different path that is valid in both $S'$ and the witness state.

*Example* 3. Returning to our Wumpus example, we show how this sampled translation reduces the size of the translation. In the example, there are four possible initial states, which we will denote by $s_{ll}, s_{lr}, s_{rl}, s_{rr}$, where $s_{ll}$ denotes the initial state in which both Wumpus are to the left of the diagonal, etc. We will select $s_{ll}, s_{lr}$ as our sampled belief state $S'$, and $s_{ll}$ as our distinguished initial state $s'$.

The set of propositions is:

- The original propositions:

    - *at-x-y* for $1 \leq x, y \leq 4$ – as we explained above, the proposition *at* is always known and does not require conditional or knowledge propositions.
    - *wumpus-at-2-3*, *wumpus-at-3-2, wumpus-at-3-4, wumpus-at-4-3*
    - *stench-at-x-y* for $1 \leq x, y \leq 4$.

- Propositions representing unconditional knowledge:

    - *Kwumpus-at-2-3*, *Kwumpus-at-3-2*, *Kwumpus-at-3-4*, *Kwumpus-at-4-3*.
    - *K¬wumpus-at-2-3*, *K¬wumpus-at-3-2*, *K¬wumpus-at-3-4*, *K¬wumpus-at-4-3*.
    - *Kstench-at-x-y* for $1 \leq x, y \leq 4$.
    - *K¬stench-at-x-y* for $1 \leq x, y \leq 4$.

- Propositions representing conditional knowledge:

    - $s_{ll}$ conditional propositions:
        * *wumpus-at-2-3/$s_{ll}$*, *wumpus-at-3-2/$s_{ll}$*, *wumpus-at-3-4/$s_{ll}$*, *wumpus-at-4-3/$s_{ll}$*.
        * *stench-x-y/$s_{ll}$* for $1 \leq x, y \leq 4$.
    - $s_{lr}$ conditional propositions:
        * *wumpus-at-2-3/$s_{lr}$*, *wumpus-at-3-2/$s_{lr}$*, *wumpus-at-3-4/$s_{lr}$*, *wumpus-at-4-3/$s_{lr}$*.
        * *stench-at-x-y/$s_{lr}$* for $1 \leq x, y \leq 4$.

- $K¬s_{ll}, K¬s_{lr}$ – propositions for denoting refuted states.

The set of actions is:

- Move: for each *move-from-$x_1$-$y_1$-to-$x_2$-$y_2$* actions we have:

    - preconditions: *at-$x_1$-$y_1$ ∧ ¬wumpus-at-$x_2$-$y_2$ ∧ K¬wumpus-at-$x_2$-$y_2$)*
    - effects [4]:
        * ¬at-$x_1$-$y_1$
        * at-$x_2$-$y_2$

- Smell: for each *smell-stench-at-x-y* action we have:

    - preconditions: *at-x-y*
    - effects:
        * *stench-at-x-y , Kstench-at-x-y*
        * *stench-at-x-y ∧ ¬stench-at-x-y/$s_{ll}$, K¬$s_{ll}$*
        * *stench-at-x-y ∧ ¬stench-at-x-y/$s_{lr}$, K¬$s_{lr}$*
        * *¬stench-at-x-y , K¬stench-at-x-y*
        * *¬stench-at-x-y ∧ stench-at-x-y/$s_{ll}$, K¬$s_{ll}$*
        * *¬stench-at-x-y ∧ stench-at-x-y/$s_{lr}$, K¬$s_{lr}$*

- Merge: we illustrate merges using the *stench-at-x-y* proposition:

    - preconditions: *(stench-at-x-y/$s_{ll}$ ∨ K¬$s_{ll}$) ∧ ( stench-at-x-y/$s_{lr}$ ∨ K¬$s_{lr}$)*
    - effects: *Kstench-at-x-y*

The initial state is a conjunction of the following conjuncts (we omit most negations for simplicity of presentation):

- The distinguished initial state:

    - *at-1-1* $\bigwedge_{x=1..4, y=1..4, x\neq 1 \vee y \neq 1} \neg$ *at-x-y*
    - *wumpus-at-2-3 ∧ wumpus-at-4-3 ∧¬ wumpus-at-3-2 ∧¬ wumpus-at-3-4*
    - *stench-at-1-3 ∧ stench-at-2-2 ∧ stench-at-2-4 ∧ stench-at-3-3 ∧ stench-at-4-2 ∧ stench-at-4-4*

- Uncertainty about the current state: $\neg K \neg s_{ll} \wedge \neg K \neg s_{lr}$.

- Conditional knowledge:

    - For the state $s_{ll}$
        * *wumpus-at-3-2/$s_{ll}$∧ ¬wumpus-at-2-3/$s_{ll}$∧ wumpus-at-4-3/$s_{ll}$∧ ¬wumpus-at-3-4/$s_{ll}$.*
        * *stench-at-1-3/$s_{ll}$∧ stench-at-2-2/$s_{ll}$∧ stench-at-2-4/$s_{ll}$∧ stench-at-3-3/$s_{ll}$∧ stench-at-4-2/$s_{ll}$∧stench-at-4-4/$s_{ll}$∧ ¬stench-at-x-y/$s_{ll}$* in all other $x-y$ locations.
    - For the state $s_{lr}$

---

4. As in this domain there are no conditional effects, we list the effect $e$ directly rather than writing $(true, e)$.

* *wumpus-at-2-3/$s_{lr}$*∧ ¬*wumpus-at-3-2/$s_{lr}$*∧ ¬*wumpus-at-3-4/$s_{lr}$*∧ *wumpus-at-4-3/$s_{lr}$*.
* *stench-at-1-3/$s_{ll}$*∧ *stench-at-2-2/$s_{ll}$*∧ *stench-at-2-4/$s_{ll}$*∧ *stench-at-3-3/$s_{ll}$*∧ *stench-at-4-4/$s_{ll}$*∧ ¬*stench-at-x-y/$s_{ll}$* in all other *x-y* locations.

Finally, the goal is K*at-4-4*.

### 5.2 A Note on Theory

The theoretical guarantees in Section 4 do not hold for the sampled translation. Specifically, as can be expected, the translation is no longer sound: a plan that works for some states may not work for all states. However, since we never apply an illegal action (because we maintain a complete description of the belief state during plan execution), if we maintain the assumption of belief-connectedness, then the goal always remains reachable. The question is whether we can ensure progress towards the goal. Unfortunately, our planner may always come up with an unsound plan, even if we have belief-connectedness, and so progress cannot be guaranteed without additional assumptions. For example, if we assume that the sample size grows each time our plan is unsound, i.e., we accumulate the witness states discussed above, then we can ensure progress is made, and that eventually the goal will be reached, i.e., we are assured completeness.

### 5.3 Belief State Maintenance through Regression

As noted above, we must maintain information about the true belief state. Belief state maintenance is a difficult task – various representations such as CNF, DNF, Prime Implicates, Prime Implicants, and more (To et al., 2009; To, Son, & Pontelli, 2010, 2011b; To et al., 2011, 2011a), all work well in some domains and poorly in other domains. That is, each representation method is suitable for a family of domains with some special features, but does not work well on domains where these features do not exist. However, we require our belief state only in order to answer two types of queries: (1) Sampling a subset of the current possible states – executed before each replanning episode for constructing the classical problem, and (2) Checking whether a literal $l$ holds in all currently possible states – executed prior to each action execution to ensure that it is not unsafe, and in order to check whether the goal has been reached.

We propose a method for answering such queries without maintaining an explicit representation of the belief space, by regressing queries through the execution history towards the initial belief state. This approach requires that we maintain the initial belief state formula and the execution history only, somewhat similarly to the situation calculus (McCarthy & Hayes, 1969; Reiter, 1991). To answer a query about the current state, we regress this query through the entire history and compare it to the initial state. To improve performance, we also cache limited information about intermediate belief states.

The main benefit of this approach is that it is focused on the query's condition only, and therefore yields small formulas for which it is easier to check satisfiability than, e.g., a formula describing the complete history in conjunction with the initial belief (Hoffmann & Brafman, 2006). However, this process must be repeated for every query.

#### 5.3.1 QUERYING FOR CURRENT STATE PROPERTIES

Throughout the online process we maintain the symbolic initial state formula and the history of actions and observations made. We use this information to check, prior to applying any action $a$, whether its preconditions hold in the current state, that is, whether the action is safe. We must also check whether the goal conditions hold in the current state to determine whether the goal has been achieved. To check whether a condition $c$ holds, we regress $\neg c$ through the current history, obtaining $\bar{c}_I$. A world state currently satisfies $\neg c$ iff it is the result of executing the current history in an initial state satisfying $\bar{c}_I$. $\bar{c}_I$ is *inconsistent* with $\varphi_I$ iff $c$ holds in all states currently possible.

More specifically, when checking whether a literal $l$ (or a set of literals) holds at the current belief state, we regress its *negation* through the current history, resulting in a formula $\psi$. Next, we check, using a SAT solver, whether $\varphi_I \wedge \psi$ is satisfiable. If it is not satisfiable, we know that $l$ is valid.

Recall that $c' = \text{regress}(c, a)$ is the weakest condition on a state $s$ such that executing $a$ in $s$ yields a state satisfying $c$. We can compute $\text{regress}(c, a)$ using the following recursive procedure:

- $\text{regress}(l, a) = \text{false}$ if $(\text{true}, \neg l)$ is in $\text{effects}(a)$.

- $\text{regress}(l, a) = \text{pre}(a)$ if $(\text{true}, l)$ is in $\text{effects}(a)$. In our case $pre(a)$ can be eliminated from the regression, because the preconditions of $a$ were already regressed and proven to be valid prior to applying $a$. Thus, if $(\text{true}, l)$ is in $\text{effects}(a)$ then $\text{regress}(l, a) = \text{true}$.

- $\text{regress}(l, a) = \text{pre}(a) \wedge (l \bigvee_{(c,l)\in\text{effects}(a)} c) \bigwedge_{(c,\neg l)} \neg c$. That is, either $l$ existed prior to the action executed, or it was added by one of the conditions that has $l$ as its effect. It is also impossible for any of the conditions removing $l$ to apply. As above, $pre(a)$ can be eliminated from the regression.

- $\text{regress}(c_1 \wedge c_2, a) = \text{regress}(c_1, a) \wedge \text{regress}(c_2, a)$

- $\text{regress}(c_1 \vee c_2, a) = \text{regress}(c_1, a) \vee \text{regress}(c_2, a)$[5]

Applying regression to histories – sequences of actions, we extend the meaning of the *regress* operator:

- $\text{regress}(c, \emptyset) = c$

- $\text{regress}(c, h \cdot a) = \text{regress}(\text{regress}(c, a), h)$

To maintain a correct description of the set of initial world states, we must also update the initial belief-state formula whenever we make an observation. Thus, we regress every obtained observation through $h$, obtaining a regressed formula $\psi$, and we conjoin $\psi$ to $b_I$. Thus, the updated set of initial state is now described by $\varphi'_I = \varphi_I \wedge \psi$. To optimize, we apply unit propagation on $\varphi'_I$ and maintain it in semi-CNF form — a conjunction of disjunctions and *xor* (*oneof*) statements. That is, convert the newly added $\psi$ to CNF form. In all the current benchmarks we find that maintaining the initial belief formula as a CNF is easy.

### 5.3.2 SAMPLING THE BELIEF STATE

SDR samples the belief state to generate a subset of possible states when computing a translation to a classical problem. Using the regression mechanism, we maintain only a formula $\varphi_I$ describing the possible set of initial states given the initial constraints on possible states and the current history, i.e., the actions that were executed and the observations that were sensed.

---

5. Recall that effects are deterministic and that the effect condition is a conjunction.

To sample $n$ states from the current belief state, we begin by finding $n$ possible satisfying assignments to the initial belief formula $\varphi_I$. We do so by running our own simple SAT solver, which picks propositions randomly from the set of unassigned propositions in the formula, sets a value to them, and propagates that value through the formula. When the formula is unsolvable, we backtrack. In all current benchmarks the structure of the initial belief formula is very simple, e.g., a disjunctive set of $oneof$ statements, or $or$ clauses, and our simple SAT solver finds solutions very rapidly.

The $n$ satisfying assignments to $\varphi_I$ represent a set of initial states that are consistent with our current information (i.e., the initial belief state and the observations made so far). To obtain a sample of states from the current belief state, we progress them through the history.

### 5.3.3 OPTIMIZATION: PARTIALLY-SPECIFIED STATES

For each step of execution of the current plan we maintain a list of literals known to hold at that belief state. All propositions that do not appear among the literals in this list are currently unknown. We call this construct a *partially-specified belief state* (PSBS), and it serves as a cache. When we execute an action $a$, we propagate this set forward and update it as follows:

- If $(c, l)$ is an effect of $a$, and $c$ must be true before $a$'s execution, we add $l$ and remove $\neg l$.

- If $l$ and $\neg l$ may both be true (that is, $l$ is unknown in the PSBS), $a$ deletes $l$ if $l$ holds (possibly conditional on other conditions that necessarily hold), and $a$ does not add $l$ when $\neg l$ (and some other conditions possible) holds, we can conclude that $\neg l$ must be true after the execution of $a$.

- While performing regression for some condition $c$, we may learn that some literal $l$ is valid in some intermediate PSBS. We update this PSBS and its successors, accordingly.

For instance, when observing a *test-passed* in Example 2, we add this to the last PSBS. Then, after the regression of the action $test_i$, we obtain $disease_i$, and add it to the previous PSBS, and so forth. Following the regression, and the simplification techniques over $\varphi'_I$ mentioned above, we may learn that some literal $l$ was true initially. We then progress $l$ through the history to add more information into PSBSs. In Example 1, for instance, when regressing a *stench* observation to the initial state formula, we may learn that *wumpus-at-3-4* holds, we then progress this through the history and add the learned information to all PSBS.

The PSBS may reduce the formulas constructed through the regression. For example, suppose our regressed formula at an intermediate belief state $b$ has the form $\varphi \wedge l$, where $l$ is a literal that belongs to the current PSBS, i.e., it is known to hold at $b$. Then, we need only regress $\varphi$ back from $b$. Or, if we know that $\neg l$ holds in $b$, we can immediately conclude that the regressed formula will be inconsistent with the initial belief state.

To conclude, our belief state representation uses the initial belief, represented as a symbolic formula (initialized as given in the PPOS definition), and sets of literals that were shown to hold for each time step. We update the initial belief by adding more formulas resulting from regressing observations, thus adding more constraints on the set of possible states, and reducing the cardinality of the initial belief state. Whenever we discover that a literal holds at any time step, either through an action (unconditional) effect, or while regressing an observation, we cache this. The regression mechanism uses these cached facts whenever possible to reduce the size of the regressed formula.

### 5.4 The Complete SDR Planner Algorithm

After explaining all the essential components of the SDR planner, i.e., sampling, translation, and regression-based belief maintenance, we can now present the complete SDR algorithm (Algorithm 2).

---

**Algorithm 2** SDR (Sampling Translation)

---

**Input:** PPOS Problem: $\mathcal{P} = \langle P, A, \varphi_I, G \rangle$, Integer: *size* – the number of states in the sampled $S'_I$

1: $h := \phi$, the empty history
2: **while** $G$ is not known at the current belief state **do**
3:    Select a distinct set of states $S'_I$ consistent with $b_I$, by finding satisfying assignments to $\varphi_I$ s.t. $|S'| \leq size$
4:    Select the distinguished state $s'_I \in S'_I$
5:    Propagate $S'_I$ and $s'_I$ through $h$, resulting in $S'$ and $s'$
6:    Generate the deterministic planning problem $\mathcal{P}_c(S', s')$
7:    Find a solution $\pi$ for $C$
8:    **if** no solution exists **then**
9:       **return** failure
10:    **end if**
11:    **while** $\pi \neq \emptyset$ **do**
12:       $a :=$ first$(\pi)$
13:       Regress $\neg pre(a)$ through $h$ obtaining $\psi_{\neg pre(a),h}$
14:       **if** $\psi_{\neg pre(a),h}$ is inconsistent with $b_I$ **then**
15:          break
16:       **end if**
17:       Execute $a$, observe $o$
18:       Append $\langle a, o \rangle$ to $h$
19:       Regress $o$ through $h$ obtaining $\psi_{o,h}$
20:       Update the initial belief state formula $\varphi_I$ given $\psi_{o,h}$: $\varphi_I \leftarrow \varphi_I \wedge \psi_{o,h}$
21:       Remove $a$ from $\pi$
22:       **if** $o$ is inconsistent with $s'$ **then**
23:          break
24:       **end if**
25:       Update the current state: $s' \leftarrow a(s')$
26:    **end while**
27: **end while**

---

The algorithm begins by querying if the goal state has already been reached (line 2). This is done by regressing the negation of the goal conditions through the current history, and checking whether the negation is consistent with the initial belief state. If the latter is true, we know that there is some state for which the goal conditions do not apply.

We then choose a sub-sample $S'_I$ of $b_I$ (the initial belief state) and $s'_I \in S'_I$ (lines 3 and 4). We propagate the states in $S'_I$ through the history by applying the executed actions in $h$ to all states in $S'_I$ (line 5). Lines 3-5 are thus equivalent to sampling $S'$ and $s'$ from the current belief state.

Once we obtain $S'$ and $s'$, we use the translation in Section 3.2 (replacing $b$ with $S'$) to generate the classical problem $\mathcal{P}_c(S', s')$, and solve it using any classical planner (lines 6 and 7). If there is no solution to this problem, then that means that the goal cannot be obtained if $s'$ is the current state, and that thus there is no solution to the PPOS.

We now execute the obtained plan; We first check if the preconditions of the current action hold in the current belief state. This is done by regressing the negation of the preconditions through the history, and checking if the regressed formula is consistent with $b_I$ (line 13-16). If it is consistent, i.e., if there is a state in $b_I$ for which the (regressed) negation of the preconditions hold, then we must choose a new $S'_I$ and replan.

Finding that the preconditions hold in the current belief state, we execute the current action, observing some observation $o$. We regress $o$ through the history and update the initial belief state using the regressed formula (lines 19 and 20). If $o$ is inconsistent with $s'$ (that is, if $\phi_{o,h}$ is inconsistent with $s'_I$) then we must sample again and replan (lines 22-24). Otherwise, even if $o$ is inconsistent with some other state $s''_I \in S'_I, s''_I \neq s'_I$, we continue executing the plan.

When the execution of the plan terminates we check again to see if the goal has been met, and if it was not obtained, we sample and replan again.

### 5.4.1 BIAS FOR OBSERVATION AND KNOWLEDGE.

It is possible to bias the SDR planner to make observations. A crude and simple method is to execute any sensing action that can sense an unknown value without affecting the state. In the context of current benchmarks this improves the planner's run-time performance. We refer to this version of SDR as SDR-Obs.

A more focused method is to augment the goal state with the requirement to "prove" that the distinguished initial state is correct. We refer to this version as SDR with state-refutation (SDR-SR). Recall that the distinguished state determines the value of all propositions in the initial state, but it does not affect our knowledge. Thus, in the Wumpus domain, with $s_{ll}$ as the distinguished state we have that *wumpus-at-2-3* holds, but K*wumpus-at-2-3* does not. If we change the goal to K*at-4-4* $\wedge K \neg s_{lr} \wedge K \neg s_{rl} \wedge K \neg s_{rr}$ then the planner will generate a plan that has the knowledge effects as well. That is, a valid plan must prove, e.g., that the initial state $s_{rr}$, where both Wumpuses are on the right, is invalid. As state refutation can be achieved only by actions whose effects are $\neg s$, and such effects are obtained only following a differentiating observation, this method encourages the planner to take sensing actions.

As the distinguished initial state is unlikely to be the true initial state, plans generated with this modified goal are likely to more quickly identify this fact. This, in turn, will cause replanning to trigger sooner, and with more information. Of course, we do not necessarily need to know the identity of the initial state to succeed, and so this may also add sensing actions that may not be required in an optimal plan.

## 6. Related Work

SDR borrows and extends ideas from various related planners, most notably: replanning/online planning, translation-based techniques, and lazy belief state representation. We briefly discuss them here.

Replanning has recently become popular for online planning under uncertainty by the FF-Replan (Yoon et al., 2007) MDP solver. Replanning is a technique for planning under uncertainty online, where at each stage, the planner solves a simplified problem that typically removes the uncertainty, e.g., by making assumptions about the value of unknown variables or the effects of actions.

The planner executes the obtained solution until it receives information that contradicts its assumptions, updates the model with the new information, and repeats the process. For example, FF-Replan assumes certain deterministic effects for stochastic actions, obtaining a classical planning problem. Because replanning essentially ignores certain aspects of the model, it runs the risk of getting stuck at dead-ends, or regions of the state space from which it is difficult to reach the goal. However, combined with smart sampling techniques recently developed for stochastic planning problems, such as UCT (Kocsis & Szepesvári, 2006) , replanning becomes a powerful technique. For example, FF-Replan was later improved using the idea of hindsight optimization (Yoon et al., 2008; Yoon, Ruml, Benton, & Do, 2010), where multiple, non-stationary determinizations of the MDP are examined. The choice of the next action is guided by the solution to the multiple resulting classical planning problems, enabling the planner to account for different possible future dynamics.

As noted above, an essential element of replanning is the reduction of the current problem into a simpler problem. SDR builds on the the translation-based approach to conformant planning introduced in Palacios and Geffner (2009) to generate a classical planning problem, specifically, their $K_{S_0}$ translation. In conformant planning, the resulting classical planning problem is equivalent to the original problem (has the same set of solutions), but may be much larger in size. Applied to contingent planning, this translation methods generates a non-deterministic, fully observable, planning problem. To make the problem deterministic, SDR simplifies it by assuming that observations conform to those of some specific initial state. To reduce its size, SDR samples a subset of the initial states.

Palacios and Geffner (2009) suggest a different technique for controlling the problem size. While SDR maintain the value of propositions conditioned on each initial state, their planner maintains the value of each proposition conditioned on *sets* of initial states, called tags. Ideally, the tags for proposition $p$ contain initial states that differ only on the value of propositions whose initial value does not affect the future value of $p$. Such sets can be quite large, implying that fewer tags are required. This, in turn, leads to considerable savings in the size of the generated problem. When the set of tags required for a complete translation is large, one may use tags that are not deterministic in the following sense: a proposition may have different values in different states belonging to this tag. While soundness can still be maintained in this case, one must sacrifice completeness.

The CLG planner (Albore et al., 2009) takes a different approach to extending the ideas of Palacios and Geffner to contingent planning. If $P$ is the original contingent planning problem, denote by $X(P)$ the fully-observable non-deterministic problem obtained by using the transformation of Palacios and Geffner. CLG solves $X(P)$ by obtaining heuristics from a relaxed version of the problem that moves each precondition as a condition in all the effects of the action in a similar way as done in CFF (Hoffmann & Brafman, 2006), drops non-deterministic effects, and introduces propositions of the form $Ap$, which roughly say that $p$ was observed. The effect of a sensing action that senses $p$ is thus $Ap$, rather than $Kp$ (which would erroneously imply that offline we know what value will be sensed). Actions that require $p$ as a precondition, require $Ap$ in the classical problem generated. This forces the classical solver to insert a sensing action that senses $p$ before it can insert an action that requires $p$. Of course, the actual value sensed online may not correspond to the assumptions made by the rest of the plan. This is why, following every sensing action, CLG (in its online version) replans. Thus, each plan is executed until the first sensing action. If the initial belief state is correctly computed, the translation ensures that this fragment of the plan can be executed. Thus, both CLG and SDR use replanning and a translation to classical planning in each replanning phase. However, SDR uses a translation that is simpler in two ways. It does not use the $Ap$ type proposi-tions, but instead uses sampling to determinize sensing actions. In addition, SDR uses sampling to select only a subset of possible initial states. This leads to much smaller classical planning problems that are faster to generate and solve, but are less informed. Consequently, often SDR takes more steps to reach the goal, but is able to scale up better than CLG.

Recently, Bonet *et. al.* introduced a replanning-based contingent planner (Bonet & Geffner, 2011) called K-planner, which is very similar to SDR, but focuses on a special class of domains that can be handled more efficiently. In these domains the hidden variables are static, i.e., the value of all hidden propositions does not change throughout the execution of the plan. For example, in the Wumpus domain, Wumpuses do not move, and the location of the Wumpuses along the diagonal is static. In the Localize domain, however, the hidden current wall configuration changes with every move action. Thus, K-planner is unsuitable for this domain. In addition, K-planner assumes that the value of observable variables is always observed following any action – i.e., there are no explicit sensing actions.

K-planner and SDR were developed in parallel, share many ideas, and provide similar theoretical guarantees. K-planner uses replanning and a translation into classical planning very much like SDR (and CLG). Whereas SDR's translation assumes that sensed value will correspond to the sampled initial state, K-planner has actions that correspond to making assumptions about the sensed values. That is, real sensing actions are translated into multiple classical actions that lead to knowledge of different values of the sensed variable. Thus, the classical planner may "choose" what value it would like the sensor to sense. This essentially allows the planner to make optimistic assumptions while ensuring goal focused sensing. However, multiple sensing actions in the plan may have "assumed" effects that cannot be realized by any initial state. This may affect the quality of the classical plan generated, but does not lead to unsound behavior because, online, the plan is executed only until the first sensing action, after which the belief state is updated and replanning takes place. SDR's use of a select initial state ensures that the sensed values are consistent, but may be more pessimistic because it cannot conceive of sensed value being different from those dictated by the special initial state selected.

As noted above, K-planner makes certain assumptions about the nature of uncertainty in the domain, and one important contribution it makes is showing how these properties can be leveraged to provide a more efficient representation and stronger guarantees. Essentially, the static hidden variable assumption made by K-planner can be viewed as saying that the uncertainty in the domain is essentially about the value of a multi-value variable, and conditional effects do not depend on the value of this variable. In that case, one can represent the belief state more compactly and one can generate very compact translations to classical planning. Bonet and Geffner prove the soundness and completeness of K-planner under similar assumptions to those made here, but they also show that under their additional assumption about the nature of uncertainty in the domain discussed above, they achieve these properties while generating classical planning problems that are linear in the size of the original planning problem.

Like most planners that operate under uncertainty, SDR maintains some form of a belief state. SDR has taken the "lazy" approach to belief state maintenance to an extreme. The lazy, or implicit approach to belief state maintenance was introduced in CFF (Hoffmann & Brafman, 2006). CFF maintains a single complete formula that describes both the history and the initial state jointly. This requires using a different copy of the state variables for each time point. An action applied at step $t$ is conjoined with this formula as clauses that describe the value of variables in time $t+1$ as a function of their value in time $t$. To determine whether condition $c$ holds at the current belief state, the current formula is conjoined with $\neg c$. If the resulting formula is consistent, we know that there are possible states in which $\neg c$ holds. Otherwise, we know that $c$ is valid. For example, if $c$ is the goal condition, we know that a plan was found. If $c$ is the precondition to some action $a$, we know that $a$ can be applied safely. CFF also caches such information discovered by simplifying the formula whenever such a conclusion is obtained, via unit propagation. The regression method that we use can be thought of as constructing, for each query, only the part of the CFF formula that is needed for answering the current query. The downside of this approach is that we could, in principle, reconstruct the same formula, or parts of a formula repeatedly. The advantage is that the formulas that we construct are much smaller and easier to satisfy than the complete CFF formula.

Many other belief state representations were explored in literature, including binary decision diagrams (Bryce, Kambhampati, & Smith, 2006), DNF and CNF representations, and Prime Implicates and Prime Implicants (To et al., 2011b, 2011, 2011a, 2009, 2010). Overall, To (2011) concludes that different domains require different representations. That is, each belief representation method works well on domains with a certain structure of actions, and not as well on other domains. It would be interesting to compare the various belief representation methods to our lazy regression based technique on a large set of benchmarks, and we leave this to future research.

To *et al.* also suggest a number of contingent planners, built using the above belief representations and an AND/OR forward search algorithm, that generate complete plan trees for contingent problems. Other planners such as CFF, POND (Bryce et al., 2006), and CLG in offline mode also compute such complete plan trees. In general, these plan trees are exponential, and must produce different plans for each possible initial state in the worst case. Thus, offline contingent planning is inherently difficult to scale up to larger domains with many possible initial states.

## 7. Experimental Results

To demonstrate the power of our replanning approach we compared SDR to the state of the art contingent planner CLG (Albore et al., 2009). We use CLG in its so-called execution mode, where it becomes an online planner. We compare SDR to CLG on domains from the CLG paper:

- Wumpus: in this grid-based navigation problem, the agent must move from the lowest left corner to the upper right corner. The diagonal of the grid is surrounded by squares containing either monsters called Wumpuses, or pits. The agent must ensure that a square is safe, i.e. no monster or pit, before entering it. Squares cannot be directly checked for safety. Instead, in squares surrounding a Wumpus the agent can smell its stench, and in squares surrounding a pit the agent can feel the breeze. The agent does not know, though, given a breeze or a stench which of the neighboring squares is unsafe. To ensure safety the agent must use information collected from various squares around the suspected squares. This domain demonstrates a complex hidden state with multiple sensing actions, but no conditional effects, which are the key bottleneck for the CLG translation. Thus, in this domain the unknown features of the environment are fixed, and the uncertainty does not propagate to other propositions , i.e. known features do not become unknown. This is the type of domains that $K$-planner can handle.

- Doors: this is again a grid-based navigation task, where the agent must move from the left of the grid to the right. Along the way there are walls with a passage (unlocked door) only in a single location. To sense whether a door is unlocked the agent must try it. The naive strategy is hence to move along the wall and try all doors until the unlocked one is found. Then, the agent has to start over with the new wall. This domain exhibits simple strategies and no conditional effects. Like the Wumpus domain, the uncertainty is not propagated here.

- Color-balls: in this domain several colored balls are hidden in a grid, and the agent must find the balls and transfer them to a bin of the appropriate color. This domain has a very large state space, and multiple sensing actions, but no conditional effects. Here, again, the propositions whose values are known cannot become unknown.

- Unix: in this domain the agent must find a file in a folder tree and transfer it to the root. There is no smart technique for searching the tree and the agent must exhaustively check every subfolder. This domain does not contain any conditional effects. In this domain too the uncertainty does not propagate.

- Localize: this is yet another grid-based navigation problem, where the agent must reach the top right corner of a grid. However, the agent is unaware of its position within the grid. The agent can sense nearby walls, and hence deduce its position within the grid. This domain presents the most complex conditional effects of all domains, and is thus the most difficult for scaling up using the CLG translation. In this domain, due to the conditional effects over unknown features, the uncertainty can propagate, i.e. known features may change their value depending on the value of variables whose value is unknown, causing them, in turn, to become unknown. For example, we can sense a wall to the right, but then move and the knowledge of observing a wall to the right is lost. This domain is unsuitable for the $K$-planner.

It is clear from the description above that the benchmark domains are somewhat limited. Specifically, most of the domains do not use conditional effects, which are challenging under partial observability. Nor are there dead-ends. We come back to these issue in Section 7.5. Nevertheless, to assess the performance of SDR in comparison to current state of the art, we evaluate it on these domains, all of which allow for scaling up using larger grids, more balls and so forth.

We implemented our SDR algorithm using *C#*. The experiments were conducted on a Windows Server 2008 machine with 24 2.66GHz cores (although each experiment uses only a single core) and 32GB RAM. We used FF (Hoffmann & Nebel, 2001) compiled under a Cygwin environment for solving the deterministic problems, and we used the Minisat SAT solver (Eén & Sörensson, 2003) to search for satisfying assignments (or lack there of). As Minisat does not provide random assignments, we also implemented a naive solver that generates random assignments for the true environment states.

### 7.1 Comparing Plan Quality and Execution Time

We begin by comparing SDR variants to CLG on a number of benchmarks. We compute the average number of actions and average time over several (25) iterations for each problem instance, where in each iteration an initial state is uniformly sampled. Arguably, computing averages in the case of contingent planning is perhaps an incorrect estimation, as averages assume a uniform sampling of conditions (initial states in our case), which is not a part of a PPOS formalism. Indeed, defining a reasonable comparison metric for online contingent planners is still an open question, but for lack of a better measure, we report measured averages.

|  | SDR | | SDR-obs | | SDR-SR | | CLG | |
|---|---|---|---|---|---|---|---|---|
| Name | #Actions | Time(secs) | #Actions | Time(secs) | #Actions | Time(secs) | #Actions | Time(secs) |
| clog huge | 82.42 (0.64) | 321.28 (9.32) | 61.17 (0.44) | 117.13 (4.19) | 82 (0.42) | 786.17 (31.19) | **51.76** (0.33) | **8.25** (0.08) |
| elog 7 | 22.08 (0.05) | 1.81 (0.02) | 21.76 (0.07) | **0.85** (0.01) | 21.52 (0.05) | 1.67 (0.02) | **20.12** (0.05) | 1.4 (0.08) |
| ebtcs 70 | **33.96** (0.8) | 17.4 (0.38) | 35.52 (0.75) | **3.18** (0.07) | 35 (0.64) | 25.18 (0.39) | 36.52 (0.86) | 73.96 (0.14) |
| CB 9-1 | 244 (6.16) | 214.1 (7.65) | 124.56 (2.49) | **71.02** (1.57) | 264.24 (5.65) | 140.64 (5.21) | **94.36** (1.83) | 129.3 (0.26) |
| CB 9-3 | 841.44 (7.23) | 912.73 (18.7) | **247.28** (2.91) | **245.87** (4.03) | 665.4 (6.37) | 565 (18.1) | 252.76 (2.66) | 819.52 (0.47) |
| CB 9-5 | 1068.6 (4.44) | 1356 (10.7) | **392.16** (2.81) | **505.48** (8.82) | 918.07 (4.72) | 716 (15.3) | PF | |
| CB 9-7 | Failed | | **487.04** (2.95) | **833.52** (15.82) | Failed | | TF | |
| doors 5 | 21.6 (0.22) | 3.76 (0.05) | 18.04 (0.18) | **2.14** (0.03) | **16.64** (0.23) | 3.05 (0.04) | **16.44** (0.18) | 2.4 (0.1) |
| doors 7 | 47.76 (0.52) | 18 (0.26) | 35.36 (0.41) | **9.29** (0.1) | 40.52 (0.45) | 17.87 (0.22) | **30.4** (0.24) | 20.44 (0.02) |
| doors 9 | 97.76 (1.1) | 72.5 (0.87) | **51.84** (0.55) | **28** (0.31) | 77.68 (0.92) | 61.57 (0.9) | **50.48** (0.5) | 38.52 (0.06) |
| doors 11 | 148.44 (1.3) | 216.52 (4.4) | 88.04 (0.91) | **79.75** (1.04) | 125.08 (1) | 174.04 (1.75) | **71.68** (0.79) | 126.59 (0.1) |
| doors 13 | 229.04 (1.7) | 524.03 (7) | 120.8 (0.93) | **158.54** (2.01) | 185.64 (1.88) | 383.88 (5.42) | **105.48** (0.89) | 330.73 (0.21) |
| doors 15 | 343.26 (2.77) | 1086 (20) | **143.24** (1.36) | **268.16** (3.78) | 252.24 (2.04) | 725.76 (8.52) | PF | |
| doors 17 | 519.13 (4.7) | 1582 (26) | **188** (1.64) | **416.88** (6.16) | 299.28 (2.78) | 1089 (17) | PF | |
| localize 3 | **8** (0.12) | 1.77 (0.03) | 8.88 (0.11) | **0.81** (0.01) | 8.88 (0.14) | 1.95 (0.02) | CSU | |
| localize 5 | 14.56 (0.24) | 7.12 (0.1) | 15.32 (0.21) | **2.87** (0.04) | **13.08** (0.22) | 8.24 (0.16) | CSU | |
| localize 9 | 28.52 (0.42) | 72.69 (1.43) | 29.44 (0.47) | **26.61** (0.48) | **22.12** (0.43) | 81.44 (1.12) | CSU | |
| localize 11 | 34.67 (0.61) | 155.6 (3.87) | 41.2 (0.83) | **77.11** (1.97) | **31.12** (0.6) | 199.35 (3.84) | PF | |
| localize 13 | 37.52 (0.62) | 396.76 (10.72) | 56.96 (0.69) | **159.53** (4.18) | **39.96** (0.54) | 387.75 (7.01) | PF | |
| localize 15 | **40.08** (0.61) | 667.22 (19.7) | 68.44 (0.9) | **352.36** (9.72) | 50.63 (0.59) | 721.53 (13.33) | PF | |
| localize 17 | **45** (0.86) | 928.56 (33.2) | 81.24 (1.16) | **527.53** (15.25) | 59.48 (0.81) | 1031 (25) | PF | |

**Table 1:** Comparing CLG (execution mode) to the various SDR methods. For domains with conditional actions (*localize*) CLG execution cannot be simulated. *TF* denotes that the CLG translation failed; *CSU* denotes that CLG cannot run a simulation with a uniform distribution; *PF* denotes that the CLG planner failed, either due to too many propositions or due to timeout.

Table 1 and Table 2 lists the results for the various SDR methods and CLG. We report results for pure SDR, SDR with observation bias (denoted SDR-obs), and SDR with state refutation added to the goal (denoted SDR-SR). For each method we report the average number of actions and the average time (seconds) until the goal is reached over 25 iterations (standard error reported in brackets). Different executions correspond in our case to different selections of initial states, because in deterministic PPOS, the initial state governs the complete observation behavior. In SDR variants, various executions also correspond to different possible samplings of states from the current belief state.

|  | SDR | | SDR-obs | | SDR-SR | | CLG | |
|---|---|---|---|---|---|---|---|---|
| Name | #Actions | Time(sec) | #Actions | Time(sec) | #Actions | Time(sec) | #Actions | Time(sec) |
| unix 1 | **9.4** (0.16) | 0.46 (0.01) | 12.2 (0.16) | 0.48 (0.01) | **9.32** (0.18) | 5.28 (0.95) | 11.68 (0.23) | **0.35** (0.01) |
| unix 2 | 31.04 (0.79) | 2.01 (0.05) | 26.44 (0.72) | **1.41** (0.03) | 29.28 (0.7) | 2.33 (0.05) | **19.88** (0.47) | 2.69 (0.01) |
| unix 3 | 78.48 (2.23) | 9.61 (0.28) | 56.32 (1.72) | **5.47** (0.18) | 100 (2.12) | 21.19 (1.04) | **51.32** (0.97) | 18.56 (0.05) |
| unix 4 | 195.8 (4.73) | 53.1 (1.38) | 151.72 (4.12) | **35.22** (0.94) | 202.24 (6.02) | 78.81 (2.38) | **90.8** (2.12) | 189.41 (0.6) |
| wumpus 5 | 27.19 (0.39) | 9.8 (0.16) | 34.72 (0.3) | 6.51 (0.07) | 26.48 (0.24) | 9.37 (0.1) | **24.12** (0.1) | **2.38** (0.09) |
| wumpus 10 | 45.18 (1.57) | 102.08 (2.17) | 70.64 (1.13) | 65.89 (1.13) | **39.72** (0.49) | 54.21 (0.72) | 40.44 (0.18) | **36.29** (0.04) |
| wumpus 15 | **61.2** (1.01) | 464.74 (10.05) | 120.14 (2.4) | **324.32** (7.14) | 72.32 (1.04) | 368.53 (12.48) | 101.12 (0.67) | 330.54 (0.25) |
| wumpus 20 | **80.33** (1.47) | 1296 (21) | 173.21 (3.4) | **773.01** (20.78) | 112.33 (3.12) | 952.52 (17.62) | 155.32 (0.95) | 1432 (0.47) |

Table 2: Comparing CLG (execution mode) to the various SDR methods. The results are averaged over 25 executions, and the standard error is reported in brackets.

|  | SDR | | SDR-obs | | SDR-SR | | CLG | |
|---|---|---|---|---|---|---|---|---|
| Name | #Actions | Time | #Actions | Time | #Actions | Time | #Actions | Time |
| logistics | 1 | 0 | 0 | 2 | 0 | 0 | 2 | 1 |
| CB | 0 | 0 | 3 | 4 | 0 | 0 | 1 | 0 |
| doors | 0 | 0 | 3 | 7 | 1 | 0 | 5 | 0 |
| localize | 0 | 0 | 0 | 7 | 6 | 0 | × | × |
| unix | 1 | 0 | 0 | 3 | 1 | 0 | 3 | 1 |
| Wumpus | 2 | 0 | 0 | 2 | 1 | 0 | 1 | 2 |
| Overall | 4 | 0 | 6 | 25 | 9 | 0 | 12 | 4 |
| Largest | 2 | 0 | 4 | 10 | 2 | 0 | 3 | 1 |

Table 3: Counting the number of times that each method performed the best in each category. The bottom row shows results for the two largest problems in each category (except for the first category where only cloghuge was considered — 11 problems overall).

In offline contingent planners that compute a complete plan tree, execution time (as opposed to planning time) is negligible. In the case of online replanning algorithms, however, execution time encapsulates various important aspects of performance, such as the belief update computation, and the time required for replanning episodes. In real applications, such as controlling robots, this would translate into the time required for the system before deciding on the next action. When this time is considerable, it is possible that the robot would stop operating for a while in order to compute its next action. Thus, execution time is an important factor in deciding which online replanning approach is most appropriate.

Execution time for CLG includes translation time for the domain, CLG execution, and plan verification time in our environment, which adds only a few seconds to each execution. The translation timeout was 20 minutes, and CLG execution was also stopped after 30 minutes. We allowed FF a timeout of 2 minutes for each replanning episode, but that timeout was never reached. In most cases FF solves the deterministic plan in a few seconds. In some domains CLG's simulator does not support uniform sampling of initial states (denoted CSU in Table 1). As in these domains SDR scales up to larger instances than CLG, this problem is not crucial to the comparison. For each domain we bolded the shortest plan and the fastest execution.

As we can see in Table 3, SDR and SDR-obs are typically faster than CLG, and the difference grows as we scale to larger instances. SDR variants also scale up to larger problems where CLG fails. In most benchmark domains, the observation bias in SDR-obs resulted in faster execution, because it is typically beneficial for SDR variants to learn as much as they can concerning the hidden state as early as possible, and replan accordingly.

In domains that all planners can solve, the efficiency (in terms of avg. steps to reach the goal) is mixed. CLG clearly generates shorter plans on Unix and Doors, but in many other instances SDR-obs is better, and on the larger Wumpus instances, SDR and SDR-SR are better. It is not surprising that in general CLG produces shorter plans, because it considers all contingencies (the complete belief state), but this is also the reason for its difficulty in scaling up to larger domains. When CLG produces plans of lesser quality, we speculate that this is because of the heuristic search in belief space embedded in CLG, not due to problems in the translation approach.

SDR also computes much smaller translated domain descriptions, ranging from 10KB to 200KB. However, a direct comparison with CLG is impossible because SDR generates parameterized domains while CLG generates grounded translations, and we hence do not provide detailed results for model sizes.

In conclusion, as expected, when one is interested in shorter plans, CLG seems to be a more appropriate candidate, but when one wishes to scale up to larger domains, SDR variants are a better choice.

### 7.2 K-Planner

K-Planner (Bonet & Geffner, 2011) is a state of the art planner that handles PPOS problems with only static hidden variables and does not use explicit sensing actions, but assumes that all possible observations are immediately available upon entering a state. In such domains the maintenance of a belief state is especially simple, and a translation is relatively easy to generate. Thus, it is no surprise that K-Planner performs much better than SDR and CLG on these domains.

Looking at Table 4 one can see that the differences are especially pronounced in the case of the Wumpus domains, where SDR-OBS has a significant overhead in updating the belief. Furthermore, K-Planner employs an optimistic heuristic, which is especially appropriate for the Wumpus domain. K-Planner simply assumes that the top-right square is safe, and goes there immediately. When the square is not safe, it traces back until it finds a passage to the top-left square, and goes there directly. In other domains, the optimistic heuristic is not as successful.

To conclude, K-Planner is by far the best approach for domains with static hidden variables, but it is difficult to see a direct extension of K-Planner that handles other types of PPOS problems.

|  | SDR-obs | | K-Planner | |
|---|---|---|---|---|
| Name | #Actions | Time(secs) | #Actions | Time(secs) |
| CB-9-1 | 124.56 ( 2.49 ) | 71.02 ( 1.57 ) | **117.04** ( 10.99 ) | **34.83** ( 3.9 ) |
| CB-9-3 | 247.28 ( 2.91 ) | 245.87 ( 4.03 ) | **219.6** ( 10.09 ) | **60.63** ( 3.05 ) |
| CB-9-5 | 392.16 ( 2.81 ) | 505.48 ( 8.82 ) | **358.08** ( 15.8 ) | **94.18** ( 3.31 ) |
| CB-9-7 | 487.04 ( 2.95 ) | 833.52 ( 15.82 ) | **458.36** ( 14.64 ) | **116.63** ( 3.24 ) |
| doors5 | 18.04 ( 0.18 ) | **2.14** ( 0.03 ) | 17 ( 1.05 ) | 4.57 ( 0.35 ) |
| doors7 | 35.36 ( 0.41 ) | **9.29** ( 0.1 ) | 33.2 ( 1.67 ) | 9.01 ( 0.55 ) |
| doors9 | **51.84** ( 0.55 ) | 28 ( 0.31 ) | 52.12 ( 2.61 ) | 15.04 ( 0.95 ) |
| doors11 | 88.04 ( 0.91 ) | 79.75 ( 1.04 ) | **80.8** ( 3.04 ) | **25.82** ( 1.2 ) |
| doors13 | 120.8 ( 0.93 ) | 158.54 ( 2.01 ) | **109.72** ( 4.76 ) | **37.96** ( 1.72 ) |
| doors15 | **143.24** ( 1.36 ) | 268.16 ( 3.78 ) | 150.88 ( 4.7 ) | **55.24** ( 2 ) |
| doors17 | **188** ( 1.64 ) | 416.88 ( 6.16 ) | 188.8 ( 5.79 ) | **79.24** ( 2.62 ) |
| unix1 | 12.2 ( 0.16 ) | **0.48** ( 0.01 ) | 9.68 ( 0.85 ) | 3.71 ( 0.25 |
| unix2 | 26.44 ( 0.72 ) | **1.41** ( 0.03 ) | 22.04 ( 2.27 ) | 8.13 ( 0.71 ) |
| unix3 | 56.32 ( 1.72 ) | **5.47** ( 0.18 ) | 45.48 ( 4.59 ) | 16.87 ( 1.56 ) |
| unix4 | 151.72 ( 4.12 ) | **35.22** ( 0.94 ) | 87.04 ( 8.54 ) | **38.81** ( 3.53 ) |
| Wumpus05 | **34.72** ( 0.3 ) | 6.51 ( 0.07 ) | 35.76 ( 1.53 ) | **2.45** ( 0.21 ) |
| Wumpus10 | **70.64** ( 1.13 ) | 65.89 ( 1.13 ) | 90.52 ( 6.4 ) | **5.39** ( 0.61 ) |
| Wumpus15 | 120.14 ( 2.4 ) | 324.32 ( 7.14 ) | **107.64** ( 4.6 ) | **7.17** ( 0.6 ) |
| Wumpus20 | 173.21 ( 3.4 ) | 773.01 ( 20.78 ) | **151.52** ( 6.29 ) | **16.03** ( 1 ) |

Table 4: Comparing SDR-OBS and K-Planner, on domains with static hidden variables.

### 7.3 Effects of $|S'_I|$

An important parameter of our algorithm is the size of $S'_I$ – the number of initial states that the deterministic planner recognizes. As this number grows, the plan must distinguish between the various states and hence becomes more accurate. However, the more states we use the larger is the translation to the deterministic problem, and the more difficult it is for FF to handle. To examine the impact of this parameter we tested the performance of SDR as a function of the size of $S'_I$. As we show in Figure 2, the plan quality (the number of actions to reach the goal) of SDR does not change considerably with the number of states. The only domain where there is a significant benefit from adding more states is Wumpus, and there one sees no farther significant improvement beyond 8 states. As expected, the running time grows with the growth in the number of states. We can conclude, hence, that, at least in current domains, there is no need to use more than a handful of states.

### 7.4 Belief Maintenance

We now examine the efficiency of our proposed lazy belief maintenance method. A natural candidate to compare against is the belief maintenance method of CFF (Hoffmann & Brafman, 2006, 2005) which introduced a lazy belief-state maintenance method motivated by the same considerations that guided us. CFF maintains a propositional formula over a sets of propositions that represent the value of each original proposition at every time point, thus, $p(t)$ represents the value of proposition $p$ at time $t$. Initially, the formula contains the formula describing the initial belief state $\varphi_I$ using propositions of the form $p(0)$. CFF updates this formula following the execution of an action. If $a$ is executed at time $t$ then certain constraints between propositions at time $t$ and $t + 1$ must be satisfied. These constraints capture the add and delete effects of $a$, as well as the frame axiom. Given this formula, to check whether some literal $l$ holds at all possible states at time $t$, CFF checks whether $l(t)$ is a consequence of the current formula. Such conclusions are cached by adding them to the formula and simplifying it using unit propagation.

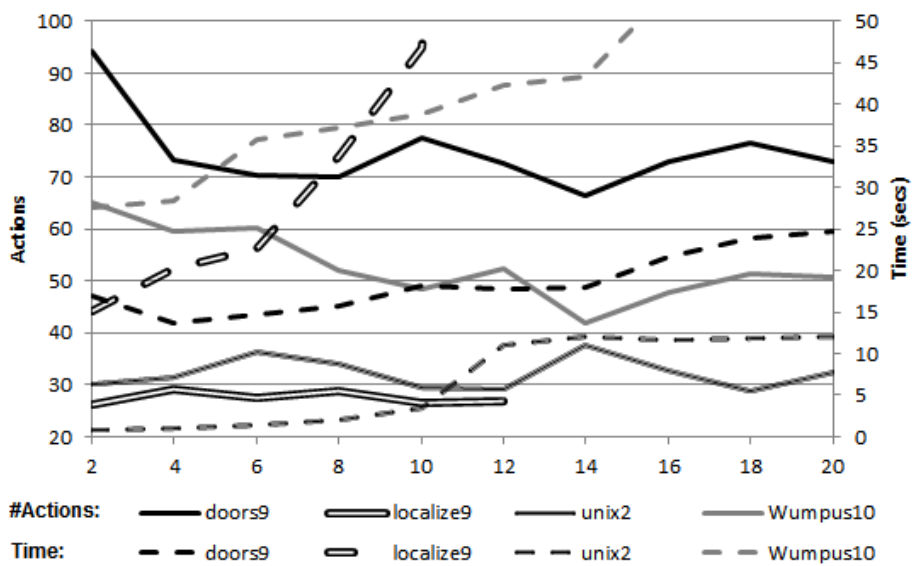

**Figure 2:** Effect of $|S'_I|$ – the number of initial states (tags) on the number of actions (solid) and the execution time (dashed) for Wumpus10, doors 9, localize 9, unix 2.

Although the update process used by CFF is quite simple, it still needs to maintain a potentially large formula and perform satisfiability checks on it. SDR's method is even lazier and reconstructs a formula for every query. This formula is focused on the literal in question, and can be much smaller but is reconstructed for each query, although as noted above, some information is cached. It is natural to ask which approach provides a better trade-off.

To compare the two methods we ran SDR once with the belief maintenance method of CFF and once with our new method. As the rest of the algorithm remains the same, and the two experiments were executed using the same random seeds, the differences in runtime stem only from the different belief maintenance method. We experimented on all the domains, but report below only the domains that the belief maintenance of CFF could solve.

As Table 5 shows, the belief maintenance method of CFF scales poorly compared to our lazy formula construction method. The only domain where the differences are not as substantial is localize, and this is mainly because in this domain the planner quickly learns the value of propositions that quickly decouple the formula into a set of formulas over disjoint sets of propositions.

To (2011) suggests a number of approaches to belief state maintenance and update. Unfortunately, while his planners are available online[6] the belief update mechanism is deeply tied into the planning mechanism, and we are currently unable to independently measure the performance of the various belief update methods and compare them to our lazy regression approach, although such a comparison is very interesting.

---

6. `http://www.cs.nmsu.edu/~sto/`

| Domain | CFF | SDR |
|---|---|---|
| cloghuge | 410.39 (4.94) | **321.28** (9.32) |
| ebtcs-70 | 481.27 (15.89) | **17.4** (0.38) |
| elog7 | 6.88 (0.94) | **1.81** (0.02) |
| doors5 | 4.28 (0.06) | **3.76** (0.05) |
| doors7 | 41.05 (1.02) | **18** (0.26) |
| doors9 | 283.93 (6.86) | **72.5** (0.87) |
| localize3 | 2.32 (0.04) | **1.77** (0.03) |
| localize5 | 7.54 (0.18) | **7.12** (0.1) |
| localize9 | 78.08 (1.69) | **72.69** (1.43) |
| localize11 | **109** (1.53) | 155.6 (3.87) |
| localize13 | 458.89 (10.06) | **396.76** (10.72) |
| localize15 | 909.12 (32.91) | **667.22** (19.7) |
| unix1 | 0.81 (0.01) | **0.46** (0.01) |
| unix2 | 6.29 (0.19) | **2.01** (0.05) |
| unix3 | 427.73 (13.22) | **9.61** (0.28) |
| Wumpus05 | 21.07 (0.27) | **9.8** (0.16) |
| Wumpus10 | 688.84 (29.46) | **102.08** (2.17) |

Table 5: Comparing the belief maintenance methods of CFF and SDR. We report the time (in seconds) of solving the domains with each of the different methods. The reported models of each type are the largest that the belief maintenance method of CFF could handle within the given timeout.

### 7.5 New Domains

We noted earlier that existing benchmark problems for contingent planning focus on problems with limited features. First, there are no dead-ends, which, as noted by Little and Thiebaux (2007), are problematic for replanning based methods. Second, the fact that we perform well with a sampled initial belief state of size 2 seems to imply that the solution is not too sensitive to the identity of the initial state. This is related to the fact that the type and amount of conditional effects we see in current domains is quite limited. Finally, the success of the sensing bias suggests that we should investigate domains in which sensing actions carry a cost, and where sensing is not necessarily separate from action. Domains where sensing actions require substantial effort to attain their preconditions may also provide interesting insights. Many of the domains above, such as colorballs, doors, and unix are also problematic because there isn't any smart exploration method that reduces the belief space faster. In all these domains the agent must move to each location independently and query for the object of interest (door, ball, or file). The agent can't, for example, sense whether the door is above or below its current position, thus cutting the belief space in half.

We thus suggest here a number of new benchmark domains, which are either variations of current domains, or adaptations of domains from the POMDP community. These new domains are especially designed to expose the difficulties of current planners, and point towards needed improvements. As such, SDR and CLG do not perform well on many of them, and sometimes fail utterly.

We first explore variations of the interesting Wumpus problem. Wumpus requires a smart exploration and sensing policy, and is thus one of the more challenging benchmarks. The original Wumpus definition requires that a cell will be "safe" before entering it. By removing this precondition, and changing the move action so that the agent is not alive if a wumpus or a pit exist in a cell, we create a domain with deadends. We experimented with this domain and, as expected, SDR and all its variations fail utterly to handle it. CLG, however, solves these domains without failing. It suc-ceeds because it fully represents the belief state and hence detects that for some possible states the agent would be dead, and would avoid such consequences. SDR, even with a complete belief representation, makes the strong assumption of a single initial state, and plans for that state specifically. As in that state the agent will not be dead in the resulting plan, it will execute it without bothering to sense for deadends. When the true world state is not the assumed state, it will eventually enter an unsafe cell and would die.

|  | SDR | SDR-OBS | SDR-SR | CLG | |
| --- | --- | --- | --- | --- | --- |
| Name |  |  |  | #Actions | Time (secs) |
| Wumpus 4 | Fail | Fail | Fail | 17.7 (0.04) | 0.17 (0.001) |
| Wumpus 8 | Fail | Fail | Fail | 40.5 (0.31) | 2.8 (0.01) |
| Wumpus 16 | Fail | Fail | Fail | 119.7 (0.91) | 182.5 (1.73) |

Table 6: Wumpus domains with deadends

We then experimented with a Wumpus variation where if the agent enters an unsafe cell, it runs back to the start cell (at 1,1). There is thus a cost for not sensing a wumpus. This cost introduces an interesting tradeoff – when the agent is close to the start cell, it is better to just enter a cell without verifying its safety. However, when the agent is far from the beginning, it is better to sense for safety rather than pay the price for going back. Here, CLG fails utterly – the underlying revised FF is unable to solve the translation within the given timeout. SDR and its variations solve this model, but we can see that the sensing bias reduces performance in this case. Still, closely looking at the plans that SDR executes, we observe that they are suboptimal – SDR does not weigh the costs of sensing vs. entering a possibly unsafe square and causing a restart. Instead, it always enters the possibly unsafe square and pays the restart price.

|  | SDR | | SDR-OBS | | SDR-SR | | CLG |
| --- | --- | --- | --- | --- | --- | --- | --- |
| Name | #Actions | Time (secs) | #Actions | Time (secs) | #Actions | Time (secs) |  |
| Wumpus 4 | **13.1** (1.13) | **3.2** (0.04) | 22.2 (0.1) | 3.4 (0.03) | 17 (0.14) | 5.4 (0.1) | Fail |
| Wumpus 8 | **34.84** (0.67) | 50.3 (1.14) | 69.12 (0.7) | 59.23 (0.74) | 37.9 (0.7) | **51.5** (1.15) | Fail |
| Wumpus 16 | **67.08** (1.1) | **178.9** (7.6) | Fail | | 74.8 (1.14) | 450.1 (14.2) | Fail |

Table 7: Wumpus domains with restarts

Next, we introduce a domain from the POMDP community, known as RockSample (Smith & Simmons, 2004), and motivated by the Mars rover task. An agent (the rover) has to sample minerals from a set of nearby visible rocks. The agent knows where the rocks are, but in order to know whether a rock should be sampled, it must activate its sensors. In our version of the problem, the agent has a sensor that senses the presence of nearby minerals, set on an antenna. When the agent extends the antenna higher, the sensor senses minerals that are farther of. When the antenna is completely folded, the agent senses only minerals in its immediate vicinity. Solving this problem smartly, without visiting rocks that do not contain minerals, requires a smart sensing strategy, involving raising and lowering the antenna in order to know which rocks contain minerals. While SDR currently solves this problem, it does not do so smartly. For example, SDR with observation bias (SDR-OBS) has no significant advantage because it does not get additional observations, as getting long-range observations requires preparation actions.

|  | SDR | | SDR-OBS | | SDR-SR | | CLG |
|---|---|---|---|---|---|---|---|
| Name | #Actions | Time (secs) | #Actions | Time (secs) | #Actions | Time (secs) |  |
| RockSample 4 | **42.2** (0.4) | 37.2 (0.36) | 45.3 (0.41) | **32.12** | 45.6 (0.39) | 37.2 (0.38) | CSU |
| RockSample 8 | **85.08** (0.65) | 109.3 109.3 (1.15) | **85.5** (0.62) | 92 (0.93) | 89.12 (0.63) | 106.04 (1.12) | CSU |
| RockSample 12 | 127.24 (0.68) | 113.4 (0.79) | 125.36 (0.81) | **101.2** (0.75) | **120.72** (0.64) | 111.66 (0.79) | CSU |
| RockSample 14 | **142.08** (0.8) | 146.75 (1.19) | 145.04 (0.63) | **128.2** (0.8) | 146.84 (0.86) | 139.3 (1.14) | CSU |

**Table 8:** RockSample domains with an $8 \times 8$ board and 4 through 14 rocks. CLG does not properly simulate the underlying world state for observations given the conditional effects and thus its performance cannot be evaluated, even though it manages to solve these domains (denoted CSU).

Finally, we also experiment with another domain that was explored in the POMDP community – the well-known MasterMind game, where the agent must guess correctly the order and color of $k$ hidden pegs out of $n$ possible colors. The agent guesses a configuration, and then receives feedback in the form of the number of correctly guessed colors and the number of correctly guessed locations. This problem is interesting because the agent never directly observes the crucial features of the state, i.e., which peg is currently correctly guessed. Furthermore, there exists an optimal sensing strategy that provably solves the game in five guesses or less for 4 pegs and 6 colors (MasterMind 6 4) (Koyama & Lai, 1993). CLG cannot solve this problem because it has a problem-width of more than 1. SDR and SDR-SR do poorly on this task because they use the simplest possible strategy – guess a setting and see if the guess was correct, i.e., whether the result was $n$ location hits, ruling out only a single state with each guess. SDR with observation bias does better, because it observes the number of location and color hits, thus ruling out many more states with each guess. It is noteworthy that the underlying FF planner does particularly badly on the translations generated for this task, executing many guesses without observing the results for no obvious reason.

|  | SDR | | SDR-OBS | | SDR-SR | | CLG |
|---|---|---|---|---|---|---|---|
| Name | #Actions | Time (secs) | #Actions | Time (secs) | #Actions | Time (secs) |  |
| MasterMind 2 4 | 25.6 (0.52) | 8.2 (0.16) | **14.48** (0.134) | **2.68** (0.02) | 28.48 (0.6) | 8.2 (0.2) | TF TF |
| MasterMind 3 4 | 63.5 (1.5) | 46.07 (1.09) | **26.76** (0.47) | **9.44** (0.12) | 66.8 (1.42) | 52.9 (1.22) | TF TF |
| MasterMind 4 6 | Fail | | **52.72** (1.08) | **74.18** (0.68) | Fail | | TF TF |

**Table 9:** MasterMind color guessing game. MasterMind $n$ $k$ stands for MasterMind with $n$ pegs and $k$ colors. CLG cannot translate this problem because it has a width above 1.

## 8. Conclusion

We described SDR, a new contingent planner that extends the replanning approach to domains with partial observability and uncertainty about the initial state. SDR also introduces a novel, lazy method for maintaining information and querying the current belief state, and has nice theoretical properties. Our empirical evaluation shows that SDR improves the state of the art on current benchmark domains, scaling up much better than CLG. However, the success of its current simple sampling techniques also highlights the weakness of current benchmark problems. To highlight this, we generated new problem domains that are challenging for current contingent planners, and can serve to measure progress in this area.

**Acknowledgments**

The authors are grateful to Alexander Albore, Hector Geffner, Blai Bonet, and Son To for their help in understanding and using their systems and to the anonymous referees for many useful suggestions and corrections. Ronen Brafman is partially supported by ISF grant 1101/07, the Paul Ivanier Center for Robotics Research and Production Management, and the Lynn and William Frankel Center for Computer Science.